\documentclass{article}
\usepackage{iclr2027_conference,times}

\usepackage[utf8]{inputenc}
\usepackage[T1]{fontenc}
\usepackage{url}
\usepackage{graphicx}
\usepackage{float}
\usepackage{wrapfig}
\usepackage{needspace}
\usepackage{placeins}
\usepackage{flafter}
\usepackage{booktabs}
\usepackage{array}
\usepackage{longtable}
\usepackage{colortbl}
\usepackage{amsmath}
\usepackage{amsfonts}
\usepackage{nicefrac}
\usepackage{microtype}
\usepackage{xcolor}
\usepackage{hyperref}
\usepackage{chunktrust_preprint}


\let\cite\citep

\definecolor{chunktrustcite}{RGB}{54,139,214}
\hypersetup{
  colorlinks=true,
  citecolor=chunktrustcite,
  pdftitle={ChunkTrust: Adapting Execution Horizons for Robot Policies with Action-Expert Evidence},
  pdfsubject={Adaptive execution horizons for robot policies},
  pdfkeywords={robotics, action chunking, adaptive horizons, vision-language-action models}
}
\definecolor{ctchunk}{HTML}{19334D}
\definecolor{cttrust}{HTML}{348447}
\newcommand{\chunktrustname}{\mbox{\textcolor{ctchunk}{Chunk}\textcolor{cttrust}{Trust}}}
\newcommand{\chunktrustabstractname}{\mbox{%
  \raisebox{-0.04em}{\includegraphics[height=0.78em]{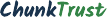}}%
}}
\definecolor{oursblue}{RGB}{226,239,251}
\definecolor{qhablue}{RGB}{226,239,251}
\definecolor{ahsyellow}{RGB}{255,246,214}
\definecolor{deltagreen}{RGB}{0,128,0}
\definecolor{deltared}{RGB}{180,0,0}
\newcommand{\ahscell}[1]{\cellcolor{ahsyellow}#1}

\newcommand{\posdelta}[1]{\textcolor{deltagreen}{\scriptsize (#1)}}
\newcommand{\negdelta}[1]{\textcolor{deltared}{\scriptsize (#1)}}
\definecolor{cttableahs}{RGB}{255,248,228}
\definecolor{cttableqha}{RGB}{233,242,250}
\definecolor{cttablegain}{RGB}{36,113,84}
\definecolor{cttableloss}{RGB}{159,66,62}
\newcommand{\ctgain}[1]{\textcolor{cttablegain}{#1}}
\newcommand{\ctloss}[1]{\textcolor{cttableloss}{#1}}

\title{\raggedright \chunktrustabstractname: Adapting Execution Horizons
for Robot Policies with \mbox{Action-Expert Evidence}}
\newcommand{\ctauthor}[2]{\mbox{\textbf{#1}\textsuperscript{#2}}}
\newcommand{\ctaffiliation}[2]{\mbox{\textsuperscript{#1}#2}}
\author{%
  \begin{minipage}{\textwidth}
  \normalfont\raggedright
  \setlength{\parskip}{0pt}
  \mbox{\ctauthor{Fanding Huang}{1,}\thanks{Equal contribution.\quad \textsuperscript{\textdagger}Corresponding authors.}}\quad
  \ctauthor{Jingyan Jiang}{3,\ensuremath{\ast}}\quad
  \ctauthor{Shifeng Bao}{2,\ensuremath{\ast}}\quad
  \ctauthor{Mingkang Pu}{1}\quad
  \ctauthor{Shiwei Li}{4}\quad
  \ctauthor{Jing Xu}{5}\quad
  \ctauthor{Shijia Xu}{6}\quad
  \ctauthor{Guanbo Huang}{1}\quad
  \ctauthor{Chenghao Gu}{1}\quad
  \ctauthor{Yuzhi Huang}{1}\quad
  \ctauthor{Chenxin Li}{7}\quad
  \ctauthor{Faisal Nadeem Khan}{1}\quad
  \ctauthor{Huan Yang}{8}\quad
  \ctauthor{Yan Wang}{1}\quad
  \ctauthor{Cheng Chi}{2,\textdagger}\quad
  \ctauthor{Zhi Wang}{1,\textdagger}\endgraf
  \vspace{0.6em}
  \ctaffiliation{1}{Tsinghua University}\quad
  \ctaffiliation{2}{Renmin University of China}\quad
  \ctaffiliation{3}{Shenzhen Technology University}\quad
  \ctaffiliation{4}{Hefei University of Technology}\quad
  \ctaffiliation{5}{Jiangnan University}\quad
  \ctaffiliation{6}{Chongqing University}\quad
  \ctaffiliation{7}{The Chinese University of Hong Kong}\quad
  \ctaffiliation{8}{Shanghai Jiao Tong University}\endgraf
  \vspace{0.6em}
  \href{https://hf618.github.io/ChunkTrust.github.io/}{\textcolor{chunktrustcite}{\textbf{Project Page}}}%
  \quad\textbar\quad
  \href{https://github.com/hf618/ChunkTrust}{\textcolor{chunktrustcite}{\textbf{GitHub}}}%
  \quad\textbar\quad
  \href{https://huggingface.co/Niugan/ChunkTrust}{\textcolor{chunktrustcite}{\textbf{Hugging Face}}}
  \end{minipage}%
}
\hypersetup{pdfauthor={Fanding Huang, Jingyan Jiang, Shifeng Bao, Mingkang Pu, Shiwei Li, Jing Xu, Shijia Xu, Guanbo Huang, Chenghao Gu, Yuzhi Huang, Chenxin Li, Faisal Nadeem Khan, Huan Yang, Yan Wang, Cheng Chi, Zhi Wang}}

\begin{document}
\maketitle

\begin{figure}[H]
  \centering
  \includegraphics[width=\textwidth]{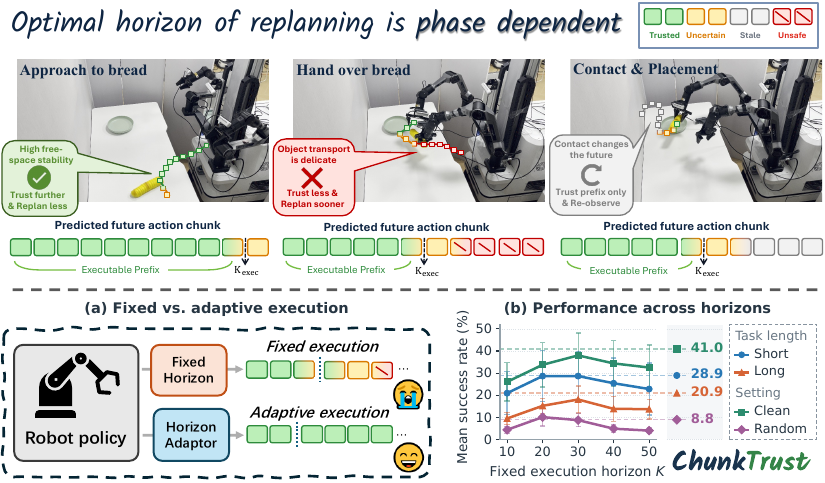}
  \caption{\textbf{Fixed execution horizons are brittle across tasks and
  settings.} \textbf{Top:} the reliable executable prefix is phase dependent in
  a real bread-manipulation rollout (qualitative trust illustration
  $K_{\mathrm{exec}}=K_t$). \textbf{(a)} A shared robot policy with
  fixed or adaptive execution horizons.
  \textbf{(b)} Fixed horizons $K\in\{10,20,30,40,50\}$ versus adaptive
  execution on $\pi_0$ RoboTwin2.0 evaluations, grouped by task length and
  environment setting. Error bars indicate $\pm 1$ standard error of the mean
  across task--setting combinations. Dashed lines show the corresponding
  adaptive references.}
  \label{fig:teaser}
\end{figure}

\begin{abstract}
Robot foundation policies predict action chunks, but how many actions to execute
before replanning depends on the current task phase. We introduce \chunktrustabstractname{},
which treats the execution horizon as a latent variable inferred from
action-expert evidence rather than a fixed hyperparameter. Its training-free
\emph{Action-aware Horizon Selector} (AHS) combines intra-chunk spectral
stability of generation traces with inter-chunk continuity between executed
history and predicted actions. An online Beta posterior with kernel forgetting
tracks horizon preferences across replans. A lightweight \emph{Query-based
Horizon Adapter} (QHA) optionally learns a context-conditioned dense prior from
complementary evidence, fused with current evidence and episode-local Beta
memory while the base policy remains frozen. Across RoboTwin2.0 and RoboCasa GR1 Tabletop, AHS improves overall
task-averaged success for each evaluated base-policy configuration, including gains of $+6.80$
percentage points on $\pi_{0.5}$ over all 50 RoboTwin2.0 tasks and $+9.67$
percentage points on Qwen3GR00T in RoboCasa. AHS+QHA raises the gain over Base
to $+9.44$ percentage points on the eight-task $\pi_{0.5}$ evaluation.
On four real-world household tasks, AHS improves the equal-task mean normalized
process score from 50.4\% to 57.5\%. Ablations examine the contributions of
both evidence terms, temporal memory, and the learned prior.

\end{abstract}

\section{Introduction}

Robot foundation policies, including Vision-Language-Action (VLA)
models~\cite{black2024pi_0,intelligence2025pi_} and World-Action Models
(WAMs)~\cite{yuan2026fast,ye2026world}, predict action chunks to amortize inference and
maintain temporal coherence. The execution horizon can differ from the generated chunk length,
reflecting a trade-off between longer execution that can preserve smooth
progress but delays correction and shorter execution that increases
feedback frequency but can disrupt coherent motion~\cite{lu2026faster}.
In manipulation, this trade-off changes within an episode: free-space approach
may tolerate longer open-loop execution, whereas contact or delicate transport demands
earlier replanning. Fig.~\ref{fig:teaser} illustrates this phase dependence in
a real rollout and shows that the best fixed horizon varies across task lengths
and evaluation settings. The resulting \emph{trust boundary} problem is to
determine how many actions from the current chunk the robot should execute
before replanning.

Adaptive chunking methods derive execution horizons from attention
structure, action entropy, or cross-horizon
agreement~\cite{wang2026vla,liang2026adaptive,jing2025mixture},
and increasingly from the policy's denoising
trajectory~\cite{feng2026dvac,chen2026geoaac}.
Other approaches learn when to replan~\cite{zhao2026dehp,xu2026bcp}
or monitor execution to trigger correction~\cite{pan2026vlacorrector}.
These approaches tackle when to replan from different perspectives,
yet an internally consistent prediction can still be
incompatible with the motion already executed.
Meanwhile, evidence at individual replans can be noisy,
while similar execution contexts recur within and across
episodes.
This raises a central question:
\emph{How can a robot identify complementary evidence native
to its action expert and internalize it as reusable knowledge
for adaptive execution?}

\begin{figure}[!t]
  \centering
  \includegraphics[width=\textwidth]{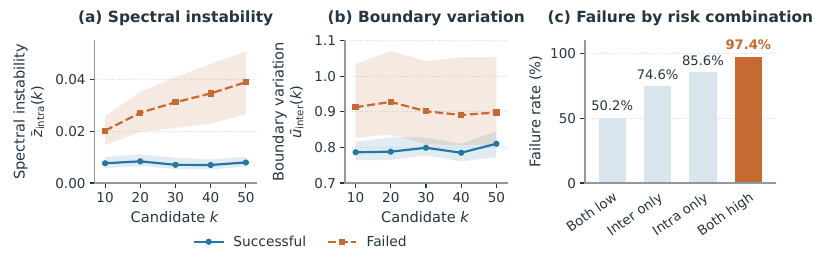}
  \caption{\textbf{Horizon-aware evidence and episode outcomes.}
  Analysis of 1,600 RoboTwin2.0 episodes with $\pi_0$ and fixed $K=H=50$.
  \textbf{(a,b)} Episode means of replan-level evidence:
  \mbox{$\bar z_{\mathrm{intra}}(k)=R_z^{-1}\sum_t z_{\mathrm{intra},t}(k)$},
  \mbox{$\bar u_{\mathrm{inter}}(k)=R_u^{-1}\sum_t u_{\mathrm{inter},t}(k)$}.
  Sums and counts $R_z,R_u$ use valid replans for each metric and horizon.
  Lines/bands show medians/interquartile ranges across episodes.
  \textbf{(c)} Failure rates after median-splitting episode risks
  (within-episode 75th percentiles of $1-q_{\mathrm{intra}}$ and
  $1-q_{\mathrm{inter}}$). ``Inter only''/``Intra only'' means only the
  named risk is high. Details: Appendix~\ref{app:fig2-diagnostics}.}
  \label{fig:metric-rationality-fixedk50}
\end{figure}

We introduce \chunktrustname{}, a framework that couples online
evidence accumulation with context-conditioned horizon
learning (Fig.~\ref{fig:method-overview}).
ChunkTrust evaluates candidate execution prefixes along two
complementary dimensions.
We assess spectral stability during action generation,
drawing on frequency-domain analyses of diffusion and flow
models~\cite{si2024freeu,huang2026exposure}.
We also assess continuity with recently executed motion,
reflecting the importance of cross-chunk consistency in
robot control~\cite{liu2025bid,black2025rtc}.
Our retrospective analysis in
Fig.~\ref{fig:metric-rationality-fixedk50} provides empirical
support for this pairing.
Failed episodes exhibit higher median spectral instability
and boundary variation, with the highest failure rate observed
when both risks are elevated.

ChunkTrust internalizes this evidence through online memory
and a learned prior.
The \emph{Action-aware Horizon Selector} (AHS) accumulates evidence
in an episode-local Beta state with forgetting, retaining
useful horizon preferences while adapting to phase changes.
The \emph{Query-based Horizon Adapter} (QHA) learns context-conditioned
horizon preferences from action-expert evidence, enabling
their reuse beyond the current episode.
At deployment, QHA supplies a dense prior that complements
online AHS evidence, combining learned preferences with
adaptation to the current rollout.

We evaluate the benefits of online horizon adaptation and
learned horizon preferences across RoboTwin2.0, RoboCasa GR1
Tabletop, and four real-world household tasks.
AHS improves aggregate success across all evaluated simulation
base-policy configurations, including gains of $6.80$ percentage
points on the complete 50-task RoboTwin2.0 suite and $9.67$
percentage points with Qwen3GR00T on RoboCasa.
Adding QHA further improves aggregate success on the eight-task
evaluation and benefits two tasks excluded from horizon-head
training, supporting reuse of the learned preferences beyond
the QHA training tasks.
On real robots, AHS increases the equal-task mean normalized
process score by $7.1$ percentage points.
Controlled comparisons and ablations examine the contributions
of complementary evidence, temporal memory, and the learned
prior, alongside alternative horizon selectors and inference
costs.

Our contributions are threefold:
(1) we formulate execution-horizon adaptation as inference over candidate
prefixes, grounded in the action expert's generation dynamics and compatibility
with executed history.
(2) we introduce AHS, which integrates dual evidence with a phase-aware Beta
posterior, and QHA, which learns a context-conditioned horizon prior 
that complements online evidence without updating the base policy.
(3) we evaluate across policy families, two simulation benchmarks, and real
robots, with full task-level results and controlled evidence, memory,
transfer, and cost analyses.

\section{Related Work}
\label{sec:related-work}

\paragraph{Action generation and reactive execution.}
Mobile ALOHA~\cite{zhaolearning} and Diffusion Policy~\cite{chi2025diffusion}
use multi-step action predictions for temporal coherence. Flow-based policies such as
$\pi_0$~\cite{black2024pi_0} and $\pi_{0.5}$~\cite{intelligence2025pi_},
and world-action models such as Fast-WAM~\cite{yuan2026fast}, extend action
generation to broader task distributions. FASTER~\cite{lu2026faster}
prioritizes near-term sampling through horizon-aware scheduling and
streaming execution. ChainVLA~\cite{huang2026chainvla} conditions successive
queries on task progress and the unexecuted action suffix.
Execution monitors offer another route to reactivity.
VLA-Corrector~\cite{pan2026vlacorrector} uses a learned latent dynamics
model to detect persistent execution drift and guide action correction.
React When You Need To~\cite{wu2026react} triggers asynchronous inference
in response to scene changes.
ChunkTrust instead selects a prefix at each replan using evidence already
available from the action expert and executed history. It neither modifies
the generated actions nor monitors new observations during that prefix.

\paragraph{Evidence-based horizon selection.}
BID~\cite{liu2025bid} selects among sampled chunks using backward coherence
and forward contrast, targeting consistency across predictions.
Horizon adaptation instead changes the executed prefix. Mixture of Horizons (MoH)
uses cross-horizon consensus~\cite{jing2025mixture}, AutoHorizon uses
action self-attention as a predictive-limit proxy~\cite{wang2026vla}, and
Adaptive Action Chunking (AAC) uses action entropy~\cite{liang2026adaptive}.
HiPolicy combines multi-frequency chunk generation with entropy-guided
execution~\cite{zhang2026hipolicy}. More recent methods expand the available
signals. Knowing When to Stop~\cite{xu2026stopping} detects entropy plateaus
in action-to-observation cross-attention, while DVAC~\cite{feng2026dvac}
measures variation in clean-action estimates during denoising.
GeoAAC~\cite{chen2026geoaac}
constructs prefix-wise geometric profiles from a single denoising trajectory.
PACE~\cite{nie2026pace} instead identifies low-speed transition points
directly in the predicted chunk.
ChunkTrust pairs spectral variation during generation with speed variation
after stitching a candidate prefix to executed history.
DVAC uses rolling history to calibrate its variance threshold, whereas
AHS maintains horizon-indexed Beta states that accumulate the paired
evidence as soft feedback with forgetting.

\paragraph{Learned horizon selection.}
DEHP~\cite{zhao2026dehp} and BCP~\cite{xu2026bcp} train horizon or
continuation heads through reinforcement learning with frozen base policies.
EQRL~\cite{wang2026eqrl} jointly learns to select the latent input,
denoising budget, and chunk length, while Spatial
Attention (SA)~\cite{park2026spatial} learns to forecast observation sensitivity
and uses it to allocate execution horizons.
QHA instead learns a context-conditioned horizon prior from complementary
action-expert evidence and combines it with current evidence and
episode-local Beta memory at deployment.
Appendix~\ref{sec:extended-related-work} discusses additional connections.

\section{Horizon-Aware Evidence from the Action Expert}
\label{sec:horizon-aware-evidence}

\subsection{Preliminaries}
\label{subsec:horizon-preliminaries}

At replan step $t$, a frozen policy conditions on visual observation $o_t$,
proprioceptive state $\mathbf x_t^{\mathrm{prop}}$, and instruction $\ell$. Its backbone produces
context tokens $\mathbf C_t=f_\theta(o_t,\mathbf x_t^{\mathrm{prop}},\ell)$, and the action expert predicts
\begin{equation}
  \hat{\mathbf A}_t =
  [\hat{\mathbf a}_{t,1},\ldots,\hat{\mathbf a}_{t,H}]
  \in \mathbb{R}^{H\times d_a}.
\end{equation}
The controller executes a prefix of length $K_t\in\{1,\ldots,H\}$ before
re-observation. We score candidates $k\in\mathcal K\subseteq\{1,\ldots,H\}$
using internal generation stability and compatibility with executed history.
The expected-round rule in Sec.~\ref{sec:method} can select intermediate
integer lengths rather than only grid points.

A single policy call with trace recording returns
$(\hat{\mathbf A}_t,\mathcal F_t)=\pi_\theta(o_t,\mathbf x_t^{\mathrm{prop}},\ell)$.
For flow-based action experts~\cite{lipmanflow}, $\mathcal F_t$ contains
velocity predictions recorded during generation, without changing the actions. We write the trace and executed history as
\begin{equation}
  \mathcal F_t=\{\mathbf v_{t,\tau}\in\mathbb{R}^{H\times d_a}\}_{\tau=0}^{T-1},
  \qquad
  \mathcal H_t=[\mathbf a^{\mathrm{exec}}_{n_t-N_t+1},\ldots,
  \mathbf a^{\mathrm{exec}}_{n_t}],
\end{equation}
where $\tau$ indexes sampling steps, $n_t$ counts actions executed before replan
$t$, and $N_t$ is the available history length. The following evidence terms use $\mathcal F_t$ and
$(\mathcal H_t,\hat{\mathbf A}_t)$, respectively.

\begin{figure}[!t]
  \centering
  \includegraphics[width=\textwidth]{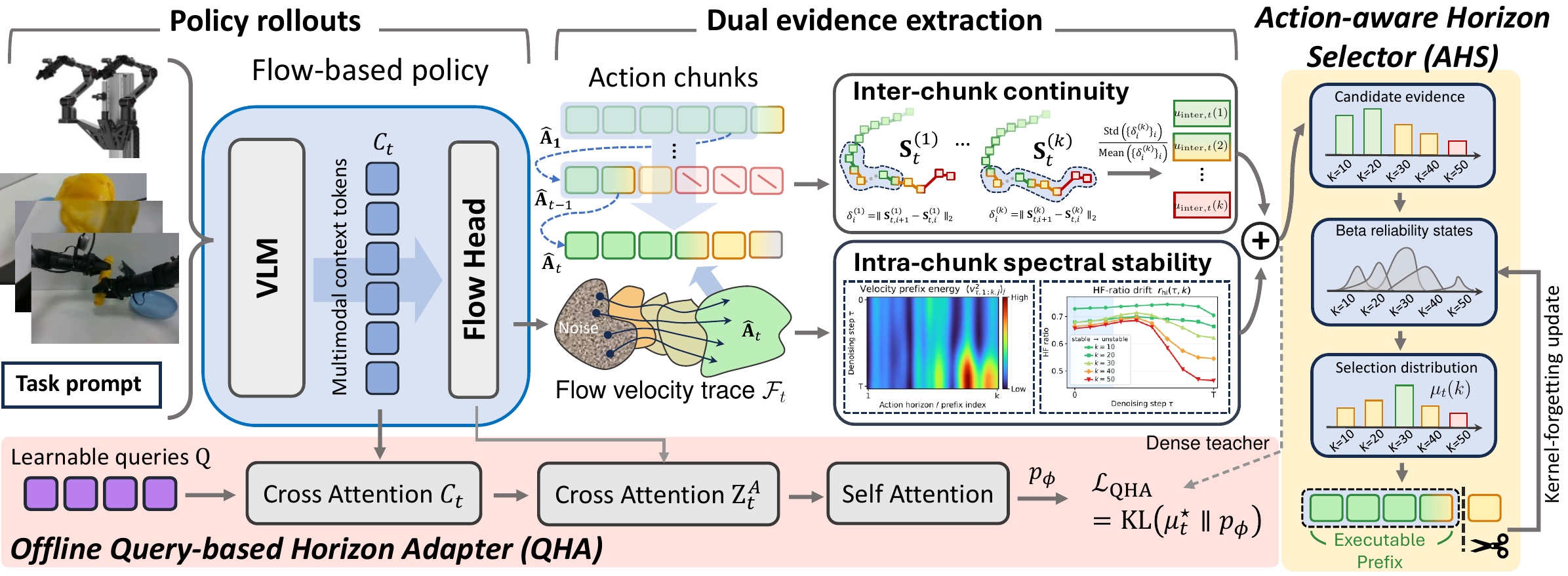}
  \caption{\textbf{Overview of horizon adaptation.}
  AHS normalizes raw evidence, maintains per-candidate Beta reliability,
  and forms selection distribution $\mu_t$.
  QHA learns a context-conditioned prior from dense evidence.}
  \label{fig:method-overview}
\end{figure}

\subsection{Intra-Chunk Spectral Stability}
\label{subsec:intra-spectral-stability}

We measure variation in the velocity-prefix spectrum during generation. 
Specifically, we apply the Fourier transform along the action horizon at each denoising step and compare the resulting spectra across steps.
For each candidate $k$, we pad its velocity prefix to length $H$ and apply a
one-dimensional Fourier transform along the action-horizon axis, separately
for action dimensions $j\in\{1,\ldots,d_a\}$ and nonnegative frequencies
$\omega\in\Omega$:
\begin{equation}
  \widehat{\mathbf V}^{(k)}_{t,\tau}(\omega,j)
  =\operatorname{FFT}_{h}\!
  \left(\bar{\mathbf v}^{(k)}_{t,\tau}[:,j]\right)_{\omega},
  \quad\text{where}\quad
  \bar{\mathbf v}^{(k)}_{t,\tau}
  =\operatorname{Pad}_{H}\!
  \left(\mathbf v_{t,\tau,1:k,:}\right)
  \in\mathbb{R}^{H\times d_a},
\end{equation}
Let $\Omega_{\mathrm{hi}}\subset\Omega$ contain frequencies above cutoff
fraction $c_{\mathrm{cut}}$ of the discrete frequency grid. The fraction of
energy in this band, aggregated across action dimensions, is
\begin{equation}
  r_{\mathrm{hi},t,\tau}(k)
  =
  \frac{\sum_{\omega\in\Omega_{\mathrm{hi}}}\sum_{j=1}^{d_a}
  \left|\widehat{\mathbf V}^{(k)}_{t,\tau}(\omega,j)\right|^2}
  {\sum_{\omega\in\Omega}\sum_{j=1}^{d_a}
  \left|\widehat{\mathbf V}^{(k)}_{t,\tau}(\omega,j)\right|^2}.
  \label{eq:hi-ratio}
\end{equation}
High-frequency energy reflects rapid variation along the future-action axis.
To track changes during generation, we define an early baseline
$\bar r_{\mathrm{hi},t,0}(k)=\frac{1}{W_\tau}\sum_{\tau=0}^{W_\tau-1}
r_{\mathrm{hi},t,\tau}(k)$ from the first $W_\tau$ sampling steps.
The raw intra-chunk instability is its root mean square (RMS) deviation over
all denoising steps, including the baseline window:
\begin{equation}
  z_{\mathrm{intra},t}(k)
  =
  \left[
  \frac{1}{T}
  \sum_{\tau=0}^{T-1}
  \left(r_{\mathrm{hi},t,\tau}(k)
  -\bar r_{\mathrm{hi},t,0}(k)\right)^2
  \right]^{1/2}.
  \label{eq:intra-raw}
\end{equation}
The early window supplies a reference rather than being discarded from the
RMS calculation. The score measures deviation from this reference, not simply
the final chunk's high-frequency energy.
Because raw scales vary across tasks, policies, and action normalizations,
we min-max normalize this score within $\mathcal K$ and set
$q_{\mathrm{intra},t}(k)=\exp[-\alpha_{\mathrm{intra}}
\tilde z_{\mathrm{intra},t}(k)]$, where $\alpha_{\mathrm{intra}}>0$ controls
the penalty strength. Larger spectral deviations thus receive lower quality.
Normalization makes this a relative comparison among candidate prefixes
at the current replan. The resulting quality need not be comparable in
absolute scale across unrelated episodes.
Fig.~\ref{fig:metric-rationality-fixedk50}(a) uses the episode mean
$\bar z_{\mathrm{intra}}(k)$ of this same replan-level score, with the exact
aggregation specified in the caption.

\subsection{Inter-Chunk Continuity}
\label{subsec:inter-continuity}

An internally stable prefix may still be incompatible with recent motion.
For continuity window $W_h$, we concatenate the available history suffix and
candidate future prefix:
\begin{equation}
  \mathbf S_t^{(k)}=
  \left[
  \operatorname{Suffix}_{(W_h-k)_+}(\mathcal H_t),
  \hat{\mathbf a}_{t,1},\ldots,\hat{\mathbf a}_{t,k}
  \right].
  \label{eq:stitched-window}
\end{equation}
Here $(x)_+=\max(0,x)$.
Let $\delta_i^{(k)}=\lVert\mathbf S_{t,i+1}^{(k)}-\mathbf S_{t,i}^{(k)}\rVert_2$
denote first-difference speed along the stitched trajectory. The raw
inter-chunk discontinuity is the speed coefficient of variation:
\begin{equation}
  u_{\mathrm{inter},t}(k)
  =\operatorname{Std}\left(\{\delta_i^{(k)}\}_i\right)/
  \operatorname{Mean}\left(\{\delta_i^{(k)}\}_i\right).
  \label{eq:inter-raw}
\end{equation}
This proxy penalizes irregular speed, including boundary jumps and stop-and-go
motion. A smaller value indicates a more uniform stitched trajectory,
whereas a larger value can reflect an abrupt correction despite a smooth
candidate viewed in isolation. As above, we min-max normalize within $\mathcal K$ and set
$q_{\mathrm{inter},t}(k)=\exp[-\beta_{\mathrm{inter}}
\tilde u_{\mathrm{inter},t}(k)]$, with $\beta_{\mathrm{inter}}>0$.
Fig.~\ref{fig:metric-rationality-fixedk50}(b) reports the corresponding
episode mean $\bar u_{\mathrm{inter}}(k)$ over valid replans.
Failed episodes have higher median raw scores for both evidence terms across
the candidate horizons. The shaded bands show interquartile ranges.

\subsection{Dual-Evidence Fusion}
\label{subsec:dual-evidence-fusion}

We combine internal stability and compatibility with recent motion into
\begin{equation}
  q_{\mathrm{mix},t}(k)=
  (1-\lambda)q_{\mathrm{intra},t}(k)
  +\lambda q_{\mathrm{inter},t}(k),
  \qquad \lambda\in[0,1].
  \label{eq:q-mix}
\end{equation}
The default is $\lambda=0.5$, with $\lambda=0$ and $\lambda=1$ recovering the
intra-only and inter-only ablations. This score supplies action-expert
evidence for horizon inference, not a deterministic execution rule or a
calibrated task-success probability.

Fig.~\ref{fig:metric-rationality-fixedk50}(c) summarizes episode risks by
the 75th percentiles of $1-q_{\mathrm{intra}}$ and $1-q_{\mathrm{inter}}$
over replans, then splits each risk at its median. Failure rises from 50.2\%
when both risks are low to 97.4\% when both are high, with intermediate rates
when only one is high. These associations motivate combining the evidence,
but they do not establish that a low-risk prefix guarantees success.
The diagnostic episodes use fixed $K=H=50$, so the comparison characterizes
the association of evidence with outcomes without selecting trajectories
based on AHS decisions. Full distributions and aggregation details
are reported in Appendix~\ref{app:fig2-diagnostics}.

\section{Action-aware Horizon Selection and Query-based Adaptation}
\label{sec:method}

AHS uses the candidate quality $q_{\mathrm{mix},t}(k)$ from
Sec.~\ref{sec:horizon-aware-evidence} to maintain an online reliability
posterior. QHA optionally learns a context-conditioned dense horizon prior
from the same evidence (Fig.~\ref{fig:method-overview}).
Neither updates the base action generator.

\subsection{Action-aware Horizon Selector}
\label{subsec:test-time-method}

\paragraph{Online reliability posterior.}
Manipulation alternates between phases such as approach, contact, transport,
and placement. A horizon that was useful in a stable phase may become
undesirable after a contact change, motivating memory that can also forget.
To retain information across noisy replans, AHS maintains a Beta state for
each $k\in\mathcal K$, initialized by $a_0(k)=b_0(k)=1$.
This is episode-local memory of evidence-derived reliability, not a supervised
task-success model. Following Thompson sampling~\cite{russo2018tutorial},
we draw a sample and combine it with the current quality:
\begin{equation}
  \hat\xi_t(k)\sim \operatorname{Beta}(a_t(k),b_t(k)),
  \qquad
  s_t(k)=\hat\xi_t(k)\,q_{\mathrm{mix},t}(k).
  \label{eq:runtime-score}
\end{equation}
Here $q_{\mathrm{mix},t}$ measures the current chunk, while sampled reliability
reflects accumulated episode-local evidence. Their product combines both.
Except for uniform exploration with probability $\epsilon_{\mathrm{exp}}$, temperature scaling and expected-round selection give:
\begin{equation}
  \mu_t(k)\propto s_t(k)^{1/T_{\mathrm{sel}}},
  \quad \sum_{k\in\mathcal K}\mu_t(k)=1,
  \qquad
  K_t =
  \operatorname{clip}_{[1,H_t^{\mathrm{avail}}]}
  \operatorname{round}\!\Bigl[\sum_{k\in\mathcal K} k\,\mu_t(k)\Bigr].
  \label{eq:expected-round}
\end{equation}
Here $H_t^{\mathrm{avail}}$ is the available chunk length. If all scores
degenerate, $\mu_t$ falls back to uniform over valid candidates.
Expected-round combines candidate preferences; exploration samples one valid
candidate uniformly. Clipping keeps the prefix within the available prediction.
AHS executes $\hat{\mathbf a}_{t,1:K_t}$, appends the executed actions to
$\mathcal H$, and replans from the next observation.

\paragraph{Posterior update with kernel forgetting.}
After execution, the soft feedback is
$y_t=\sum_{k\in\mathcal K}\mu_t(k)q_{\mathrm{mix},t}(k)$, or the sampled
candidate quality on exploration steps.
A Gaussian kernel $w_t(k)\propto\exp[-(k-K_t)^2/(2\sigma_K^2)]$,
normalized over $\mathcal K$, shares feedback among nearby horizons.
Exponential forgetting~\cite{raj2017tuning} updates the state:
\begin{align}
  a_{t+1}(k) &= \rho_f a_t(k)+\eta\,w_t(k)y_t,\\
  b_{t+1}(k) &= \rho_f b_t(k)+\eta\,w_t(k)(1-y_t),
  \label{eq:kernel-forget-update}
\end{align}
where $\rho_f$, $\eta$, and $\sigma_K$ control forgetting, update strength,
and neighborhood sharing. Decaying old evidence lets the selector adapt
as the task changes phase.
The feedback comes from action-expert quality, not an observed task-success
label. Neighboring candidates receive shared soft evidence rather than
independent rollout outcomes.
Kernel sharing couples nearby lengths; forgetting discounts earlier phases.

\subsection{Training-Time Query-Based Horizon Adapter}
\label{subsec:training-time-method}

\begin{table}[!t]
\centering
\caption{\textbf{AHS performance across simulation benchmarks.}
Success rate (\%). Each comparison uses the same policy checkpoint.
$\Delta=\mathrm{SR}_{\mathrm{AHS}}-\mathrm{SR}_{\mathrm{Base}}$.}
\label{tab:ahs-summary}
\begingroup
\fontsize{10}{11.5}\selectfont
\setlength{\tabcolsep}{3pt}
\renewcommand{\arraystretch}{1.0}
\begin{tabular*}{0.90\linewidth}{@{\extracolsep{\fill}}lll r>{\columncolor{cttableahs}}r r@{}}
\toprule
Policy & Task Scope & Train Recipe & Base & +AHS & $\Delta$ (pp)\\
\midrule
\multicolumn{6}{@{}l}{\textbf{RoboTwin2.0}\quad\normalfont Easy and Hard}\\
\addlinespace[2pt]
$\pi_{0.5}$ & 50 tasks & multitask post-training & 56.70 & \textbf{63.50} & \ctgain{+6.80}\\
$\pi_0$ & 8 tasks & task-specific post-training & 18.19 & \textbf{21.38} & \ctgain{+3.19}\\
$\pi_{0.5}$ & 8 tasks & task-specific post-training & 29.63 & \textbf{36.25} & \ctgain{+6.63}\\
Fast-WAM & 8 tasks & multitask post-training & 86.81 & \textbf{88.13} & \ctgain{+1.31}\\
\midrule
\multicolumn{6}{@{}l}{\textbf{RoboCasa GR1 Tabletop}}\\
\addlinespace[2pt]
$\pi_{0.5}$ & 24 tasks & multitask post-training & 40.08 & \textbf{42.50} & \ctgain{+2.42}\\
GR00T N1.5 & 24 tasks & zero-shot & 43.25 & \textbf{44.92} & \ctgain{+1.67}\\
GR00T N1.6 & 24 tasks & zero-shot & 47.61 & \textbf{51.42} & \ctgain{+3.80}\\
Qwen3GR00T & 24 tasks & multitask post-training & 47.83 & \textbf{57.50} & \ctgain{+9.67}\\
\bottomrule
\end{tabular*}\par
\endgroup
\end{table}

QHA predicts a dense distribution
$p_\phi(K_t=h\mid\mathbf C_t,\mathbf Z_t^A)$ for $h\in\{1,\ldots,H\}$
from VLM context tokens $\mathbf C_t$ and action-latent tokens $\mathbf Z_t^A$.
With a sparse candidate grid, AHS explicitly scores each candidate, while
QHA predicts a preference for every action step in one forward pass.
Learned horizon queries use bridge cross-attention over both token streams
(Fig.~\ref{fig:method-overview}). Architecture and teacher construction are
specified in Appendix~\ref{sec:qha-details}. The teacher normalizes dense
evidence without online Beta memory:
\begin{equation}
  \mu_t^\star(h)\propto q_{\mathrm{mix},t}(h)^{1/T_{\mathrm{teach}}},\qquad
  \sum_{h=1}^{H}\mu_t^\star(h)=1,\qquad
  h=1,\ldots,H.
  \label{eq:teacher-distribution}
\end{equation}
Only QHA is trained, minimizing
$\mathcal L_{\mathrm{QHA}}=\operatorname{KL}(\mu_t^\star\,\Vert\,p_\phi)$.
QHA thus encodes horizon preferences across training episodes in its parameters,
providing cross-episode memory that complements AHS's episode-local state.
At deployment, let $\tilde q_{\mathrm{mix},t}(h)$ denote quality on the dense
grid, interpolated from sparse evidence or scored directly on that grid.
QHA supplies a prior, combined with dense Beta reliability samples as
\begin{equation}
  s_{\mathrm{eff},t}(h)=
  \hat\xi_t(h)\,\tilde q_{\mathrm{mix},t}(h)\,p_\phi(K_t=h\mid\mathbf C_t,\mathbf Z_t^A)^\gamma,
  \qquad h=1,\ldots,H,
  \label{eq:qha-prior-main}
\end{equation}
where $\gamma\geq0$ controls prior strength. Applying temperature scaling
and the selection rule above to this dense score yields AHS+QHA, combining
learned preferences with current episode evidence.
The dense prior supplies a context-conditioned preference before the
episode-local state has accumulated much evidence. It does not remove the
need to record generation traces or evaluate AHS candidates in the hybrid
setting, as those computations provide the online correction to the prior.

\section{Experiments}
\label{sec:experiments}

\begin{table}[!t]
\centering
\caption{\textbf{QHA augmentation and transfer on RoboTwin2.0.}
Success rate (\%). Easy and Hard denote clean and randomized settings.
\textbf{A:} Evaluation on the eight tasks used to train QHA.
\textbf{B:} For $\pi_{0.5}$, QHA is trained on six tasks and evaluated on two held-out tasks.
Results are averaged over Easy and Hard.
Bold marks the best method for each policy and evaluation condition.}
\label{tab:qha-transfer}
\begingroup
\fontsize{9}{10.8}\selectfont
\setlength{\tabcolsep}{1pt}
\renewcommand{\arraystretch}{1.0}
\begin{tabular*}{\linewidth}{@{\extracolsep{\fill}}lrr>{\columncolor{cttableahs}}r>{\columncolor{cttableahs}}r>{\columncolor{cttableqha}}r>{\columncolor{cttableqha}}rrr>{\columncolor{cttableahs}}r>{\columncolor{cttableahs}}r>{\columncolor{cttableqha}}r>{\columncolor{cttableqha}}r@{}}
\toprule
\multicolumn{13}{@{}l}{\textbf{A. QHA training-task evaluation}}\\
\addlinespace[2pt]
& \multicolumn{6}{c}{$\pi_0$~{\footnotesize\citep{black2024pi_0}}} & \multicolumn{6}{c}{$\pi_{0.5}$~{\footnotesize\citep{intelligence2025pi_}}}\\
\cmidrule(lr){2-7}\cmidrule(lr){8-13}
Task & \multicolumn{2}{c}{Base} & \multicolumn{2}{c}{AHS} & \multicolumn{2}{c}{AHS+QHA}
     & \multicolumn{2}{c}{Base} & \multicolumn{2}{c}{AHS} & \multicolumn{2}{c}{AHS+QHA}\\
& Easy & Hard & Easy & Hard & Easy & Hard & Easy & Hard & Easy & Hard & Easy & Hard\\
\midrule
Blocks Ranking RGB & 19.00 & 0.00 & 28.00 & 1.00 & \textbf{31.00} & \textbf{5.00} & 36.00 & 20.00 & 48.00 & 29.00 & \textbf{56.00} & \textbf{33.00} \\
Handover Block & 41.00 & \textbf{10.00} & 41.00 & 5.00 & \textbf{64.00} & 9.00 & \textbf{44.00} & \textbf{14.00} & \textbf{44.00} & \textbf{14.00} & 32.00 & 10.00 \\
Handover Mic & \textbf{100.00} & 2.00 & \textbf{100.00} & 24.00 & \textbf{100.00} & \textbf{30.00} & 98.00 & \textbf{64.00} & \textbf{100.00} & 56.00 & 98.00 & 59.00 \\
Hanging Mug & 17.00 & 3.00 & 14.00 & 9.00 & \textbf{23.00} & \textbf{10.00} & \textbf{14.00} & 8.00 & 13.00 & 12.00 & 13.00 & \textbf{14.00} \\
Place A2B Left & 24.00 & \textbf{1.00} & 34.00 & 0.00 & \textbf{39.00} & \textbf{1.00} & 41.00 & 4.00 & \textbf{47.00} & 4.00 & \textbf{47.00} & \textbf{15.00} \\
Place Bread Basket & 11.00 & 8.00 & \textbf{18.00} & \textbf{16.00} & 15.00 & 10.00 & 30.00 & 17.00 & 50.00 & 29.00 & \textbf{55.00} & \textbf{41.00} \\
Place Bread Skillet & 15.00 & 2.00 & \textbf{23.00} & \textbf{5.00} & \textbf{23.00} & 2.00 & 24.00 & 10.00 & 36.00 & \textbf{19.00} & \textbf{44.00} & \textbf{19.00} \\
Place Can Basket & \textbf{33.00} & \textbf{5.00} & 22.00 & 2.00 & \textbf{33.00} & 3.00 & 35.00 & 15.00 & 50.00 & 29.00 & \textbf{55.00} & \textbf{34.00} \\
\midrule
Mean by setting & 32.50 & 3.88 & 35.00 & 7.75 & \textbf{41.00} & \textbf{8.75} & 40.25 & 19.00 & 48.50 & 24.00 & \textbf{50.00} & \textbf{28.13} \\
\textbf{Overall} & \multicolumn{2}{c}{18.19} & \multicolumn{2}{>{\columncolor{cttableahs}}c}{21.38} & \multicolumn{2}{>{\columncolor{cttableqha}}c}{\textbf{24.88}} & \multicolumn{2}{c}{29.63} & \multicolumn{2}{>{\columncolor{cttableahs}}c}{36.25} & \multicolumn{2}{>{\columncolor{cttableqha}}c}{\textbf{39.06}} \\
\bottomrule
\end{tabular*}\par
\endgroup
\vspace{6pt}
\begingroup
\fontsize{9}{10.8}\selectfont
\setlength{\tabcolsep}{4pt}
\renewcommand{\arraystretch}{1.0}
\begin{tabular*}{\linewidth}{@{}p{\dimexpr\linewidth-257pt\relax}>{\raggedleft\arraybackslash}p{46pt}>{\columncolor{cttableahs}\raggedleft\arraybackslash}p{46pt}>{\columncolor{cttableqha}\raggedleft\arraybackslash}p{62pt}>{\raggedleft\arraybackslash}p{71pt}@{}}
\toprule
\multicolumn{5}{@{}l}{\textbf{B. QHA transfer to held-out tasks ($\pi_{0.5}$)}}\\
\addlinespace[2pt]
Task & Base & AHS & AHS+QHA & $\Delta$ vs. AHS (pp)\\
\midrule
Blocks Ranking RGB & 28.00 & 38.50 & \textbf{43.50} & \ctgain{+5.00} \\
Place Bread Basket & 23.50 & 41.50 & \textbf{42.00} & \ctgain{+0.50} \\
\midrule
\textbf{Overall} & 25.75 & 40.00 & \textbf{42.75} & \ctgain{+2.75} \\
\bottomrule
\end{tabular*}\par
\endgroup
\end{table}

We evaluate on RoboTwin2.0~\cite{chen2025robotwin}, RoboCasa GR1
Tabletop~\cite{nasiriany2024robocasa}, and real robots. Each Base--AHS
comparison fixes the policy checkpoint. AHS changes execution without updating
policy weights. Simulation tables report success rates (\%). In
Tabs.~\ref{tab:ahs-summary} and~\ref{tab:qha-transfer}, tasks are weighted
equally after averaging Easy and Hard within each RoboTwin2.0 task.
Rounding follows aggregation. Easy and Hard denote clean and randomized
settings. Appendix~\ref{sec:exp-setup} specifies checkpoints, candidate
horizons, and evaluation protocols.

\subsection{AHS across Simulation Benchmarks}
\label{sec:ahs-benchmarks}

Tab.~\ref{tab:ahs-summary} tests AHS across benchmarks and training regimes.
Its task scope and training recipe distinguish evaluation coverage from
checkpoint provenance. Absolute scores across regimes are not matched-policy
comparisons.
The evaluation spans task-specific post-training, multitask post-training,
and zero-shot deployment on the target benchmark. Gains in these regimes
test whether execution adaptation remains useful across the evaluated
backbones. They do not imply that AHS repairs every failed task or replaces
policy training.

\paragraph{RoboTwin2.0.}
The full-suite evaluation uses one multitask-post-trained
$\pi_{0.5}$~\cite{intelligence2025pi_} checkpoint on 50 tasks, with
20 rollouts per task and setting. AHS improves success from 56.70\% to
63.50\% ($+6.80$ percentage points). The eight-task evaluation uses
task-specific $\pi_0$~\cite{black2024pi_0} and $\pi_{0.5}$ checkpoints
and multitask Fast-WAM~\cite{yuan2026fast}, with 100 rollouts per task and
setting. All three improve in aggregate. These are distinct checkpoint and
evaluation cohorts, not subset and full-suite results for one policy.
Fast-WAM improves from 86.81\% to 88.13\%, showing that execution adaptation
can still help a stronger baseline, although its aggregate gain is smaller
than those of the task-specific policies.
Appendix Tabs.~\ref{tab:robotwin50-full} and~\ref{tab:robotwin-additional}
retain full per-task outcomes, including regressions.

\paragraph{RoboCasa GR1 Tabletop.}
We evaluate all 24 tasks; the $\pi_{0.5}$ control uses 50 rollouts per task.
The multitask-post-trained $\pi_{0.5}$ and
Qwen3GR00T~\cite{ye2026starvla,community2026starvla} improve from
40.08\% to 42.50\% and from 47.83\% to 57.50\%, respectively.
Both zero-shot Isaac-GR00T baselines~\cite{bjorck2025gr00t} also improve
in aggregate. All four use $\mathcal K=\{4,8,12,16\}$.
Appendix Tab.~\ref{tab:robocasa-full} provides the complete breakdown.
Fixed-horizon, action-only, temporal-memory, and AAC~\cite{liang2026adaptive}
comparisons appear in Appendix~\ref{sec:horizon-selector-comparisons}.

\subsection{QHA Augmentation and Transfer}
\label{sec:qha-evaluation}

\paragraph{Augmentation on QHA training tasks.}
Each policy family uses one QHA head trained on eight tasks with the
base policies frozen (Tab.~\ref{tab:qha-transfer}A).
AHS+QHA improves overall success over AHS from 21.38\% to 24.88\%
for $\pi_0$ and from 36.25\% to 39.06\% for $\pi_{0.5}$.
Both Easy and Hard means improve, but $\pi_{0.5}$ regresses on Handover Block
in both settings. This may reflect a mismatch between the shared horizon
prior and the feedback timing required during bimanual object transfer.

\paragraph{Transfer to tasks held out from QHA training.}
A separate $\pi_{0.5}$ head trains on six tasks and tests on two held-out
tasks (Tab.~\ref{tab:qha-transfer}B). AHS+QHA improves the equal-task mean
from 40.00\% to 42.75\%. Blocks Ranking RGB improves from 38.50\%
to 43.50\%, compared with 41.50\% to 42.00\% on Place Bread Basket. Only the horizon head is task-held-out:
base policies remain task-specific. Appendix~\ref{sec:qha-heldout}
gives the split, per-setting results, QHA-only control, and decision agreement.

\subsection{Real-World Deployment}

\begin{figure}[!t]
  \centering
  \begin{minipage}[t]{0.65\textwidth}
    \centering
    \textbf{(a) Real-world manipulation tasks}\\[0.35em]
    \includegraphics[width=\linewidth]{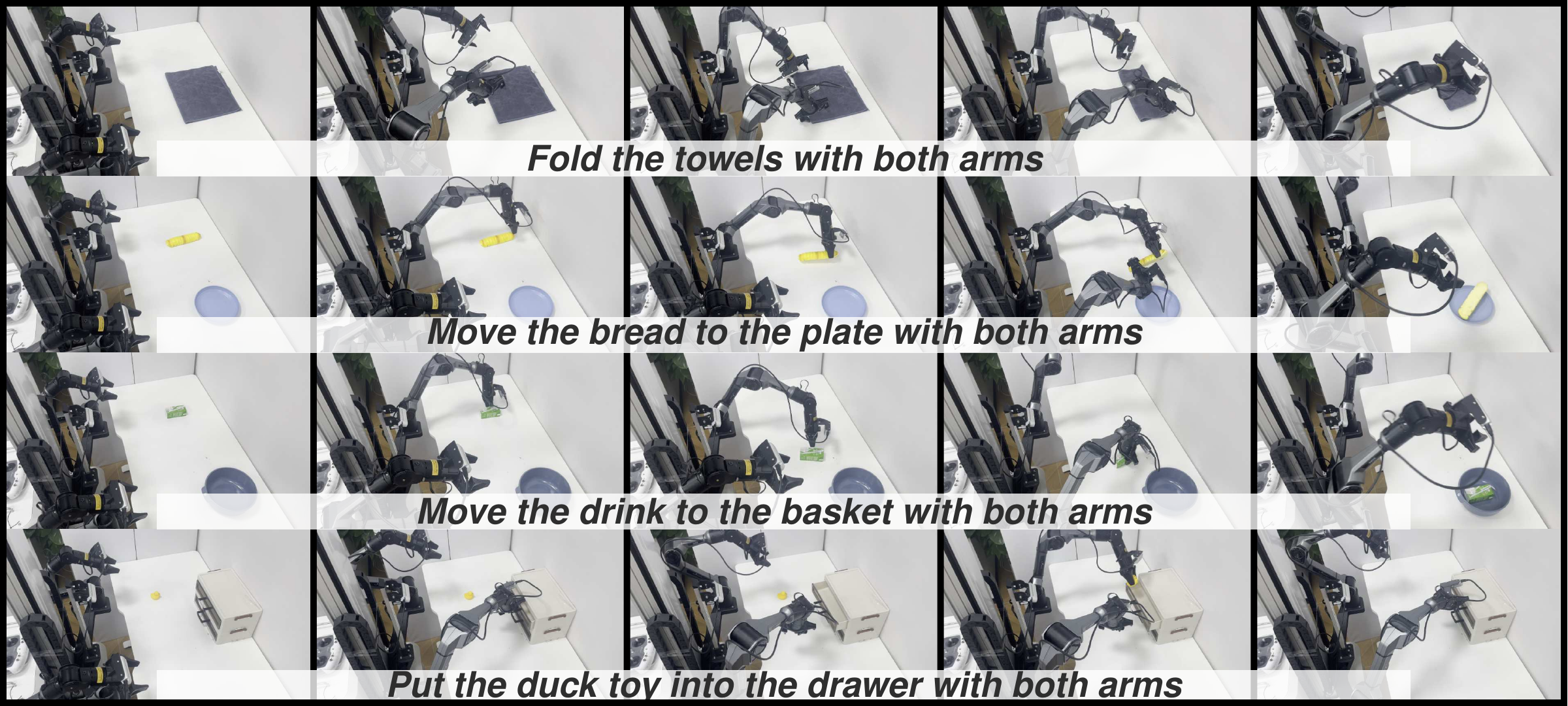}
  \end{minipage}\hfill
  \begin{minipage}[t]{0.34\textwidth}
    \centering
    \textbf{(b) Real-world performance}\\[0.35em]
    \includegraphics[width=\linewidth]{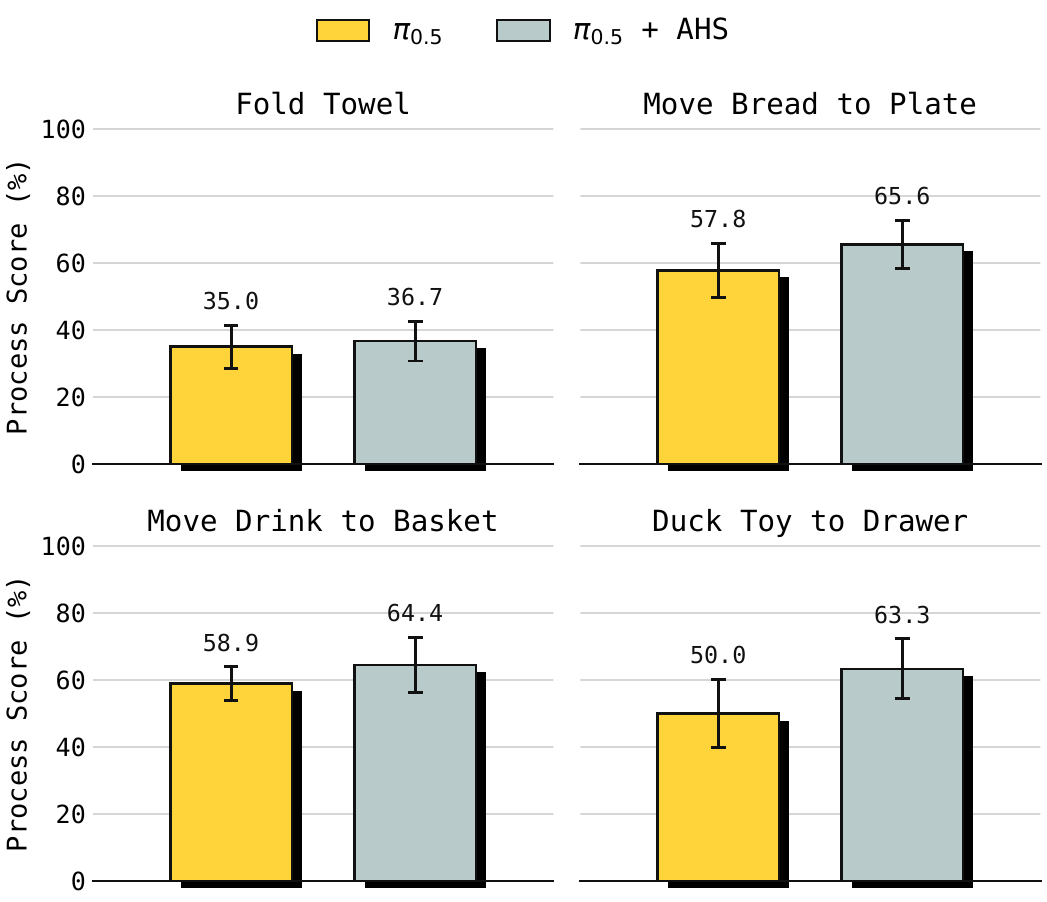}
  \end{minipage}
  \caption{\textbf{Real-world deployment.}
  \textbf{(a)} Four bimanual household tasks: fold towels, move bread to a
  plate, move a drink to a basket, and put a duck toy into a drawer.
  \textbf{(b)} Mean normalized process scores for $\pi_{0.5}$ and
  $\pi_{0.5}$ + AHS over 15 rollouts per task and method.
  Error bars show $\pm$ one standard error of the mean across rollouts.}
  \label{fig:real-robot-tasks}
\end{figure}

We deploy $\pi_{0.5}$ on the AgileX COBOT Magic ALOHA-style bimanual platform,
comparing fixed $K=25$ with AHS over 15 rollouts on each of four
household tasks (Fig.~\ref{fig:real-robot-tasks}). Object positions vary across
rollouts to test spatial generalization. Scores average predefined
sub-steps and rollouts to measure partial progress, rather than binary
success (Appendix~\ref{sec:real-world-setup}). AHS raises the equal-task
mean from 50.4\% to 57.5\%, with the largest gain on Duck Toy to Drawer: 50.0\% to 63.3\%.

\subsection{Ablation Studies}

\begin{table}[!t]
  \centering
  \small
  \caption{\textbf{Candidate-set scaling on $\pi_{0.5}$.}
  Success and per-replan overhead over four RoboTwin2.0 tasks as
  $|\mathcal K|$ increases. Fixed denotes $\pi_{0.5}$ without horizon
  selection. $\Delta$Success is in percentage points.}
  \label{tab:pi05-qmix-overhead}
  \setlength{\tabcolsep}{3.5pt}
  \resizebox{\textwidth}{!}{%
  \begin{tabular}{@{}lccccccc@{}}
    \toprule
    \textbf{Candidate set} & \textbf{$|\mathcal K|$} & \textbf{Success (\%)} & \textbf{$\Delta$Success} & \textbf{Replan time (ms)} & \textbf{Overhead (\%)} & \textbf{Replan counts} & \textbf{Ep.\ time (s)} \\
    \midrule
    Fixed               & --  & 22.00 & --      & 95.83 & --     & 4.22 & 22.43 \\
    $\mathcal K_5$       & 5   & 33.00 & +11.00  & 96.50 & 0.70   & 8.36 & 23.17 \\
    $\mathcal K_{10}$    & 10  & 32.50 & +10.50  & 96.84 & 1.05   & 9.34 & 23.08 \\
    $\mathcal K_{50}$    & 50  & \textbf{34.88} & \textbf{+12.88} & 99.28 & 3.60 & 10.57 & \textbf{21.39} \\
    \bottomrule
  \end{tabular}
  }
\end{table}

\begin{table}[!t]
  \centering
  \caption{\textbf{Component ablation ($\pi_0$).}
  Four-task mean success. Top two: bold/underline.}
  \label{tab:pi0-ablation}
  \setlength{\tabcolsep}{5pt}
  \renewcommand{\arraystretch}{1.05}
  \begin{tabular}{@{}lccc@{}}
    \toprule
    \textbf{Method} & \textbf{Easy (\%)} & \textbf{Hard (\%)} & \textbf{Overall (\%)} \\
    \midrule
    Base $\pi_0$ (default)         & 20.75 & 4.00 & 12.38 \\
    Inter-chunk only               & 23.50 & 4.00 & 13.75 \\
    Intra-chunk only               & 23.75 & \underline{5.25} & 14.50 \\
    Both evidence terms, no posterior & \underline{25.50} & 3.75 & 14.63 \\
    Full AHS                       & 24.25 & \textbf{5.75} & \underline{15.00} \\
    Full AHS+QHA                   & \textbf{27.50} & 4.00 & \textbf{15.75} \\
    \bottomrule
  \end{tabular}
\end{table}

All three studies use Place A2B Left, Place Bread Basket, Place Bread
Skillet, and Place Can Basket, with 100 rollouts per task and condition.
All three studies include Easy and Hard.

\paragraph{Candidate-set scaling.}
On $\pi_{0.5}$, five candidates raise success from 22.0\% with fixed
$K=50$ to 33.0\%, versus 34.9\% with 50 candidates
(Tab.~\ref{tab:pi05-qmix-overhead}).
Added per-replan cost rises from 0.67 to 3.45\,ms, or 0.70--3.60\%
of policy inference time. Sparse candidates thus capture most of the gain.
The table's episode times cover successes only, so they do not establish
all-attempt cost. Appendix~\ref{sec:runtime-costs} reports all-episode
costs for additional cohorts. Appendix~\ref{sec:ahs-sensitivity} gives
hyperparameter sweeps.

\paragraph{Component ablation.}
On $\pi_0$, combining both evidence terms without memory gives 14.6\%
overall but 3.75\% on Hard (Tab.~\ref{tab:pi0-ablation}). Full AHS
reaches 15.0\% overall and the best Hard result, 5.75\%, supporting
temporal memory. QHA raises overall success to 15.8\% while Hard falls
to 4.00\%. Appendix~\ref{sec:horizon-selector-comparisons} compares
memory designs with shared evidence and candidates.

\begin{figure}[!t]
  \centering
  \includegraphics[width=\linewidth]{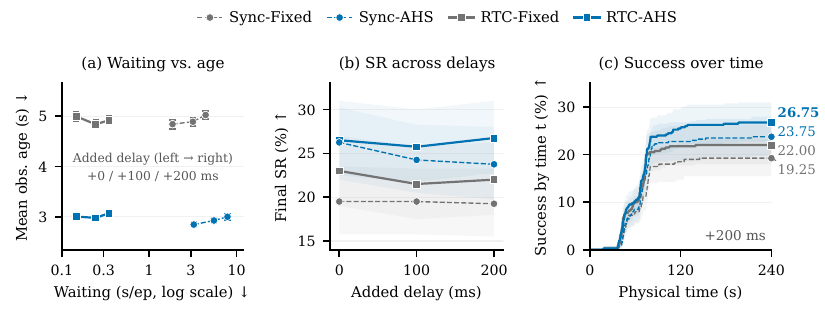}
  \caption{\textbf{AHS with asynchronous execution on $\pi_{0.5}$ Hard.}
  (a) Mean waiting and observation age. Points run left to right as
  +0/+100/+200\,ms additional delay. (b) Final SR. (c) SR($t$) at +200\,ms.
  All 400 outcomes per condition are included. Bars and shading are
  marginal/pointwise 95\% paired-seed bootstrap intervals within four tasks.
  Time uses a controlled physical clock with a 142\,ms base delay,
  not native deployment timing. Easy curves and full results:
  Appendix~\ref{sec:rtc-latency-audit}.}
  \label{fig:rtc-latency}
\end{figure}

\paragraph{Latency and asynchronous execution.}
We combine AHS with real-time chunking (RTC; \citealp{black2025rtc})
on task-specific $\pi_{0.5}$ policies (Fig.~\ref{fig:rtc-latency}).
All predict $H=50$ actions. Fixed uses target budget $K=40$, and AHS
scores candidates in $\{10,20,30,40\}$ without QHA. At +200\,ms, RTC reduces
AHS waiting from 7.893 to 0.344\,s/episode on Hard and from 6.310 to
0.344\,s on Easy. Within RTC, AHS lowers mean observation age from
4.92 to 3.07\,s on Hard and from 5.00 to 3.18\,s on Easy, while SR
rises from 22.00\% to 26.75\% and from 39.00\% to 42.75\%, respectively.
This requires more policy calls and model computation
(Appendix~\ref{sec:rtc-latency-audit}). RTC does not uniformly improve
SR over synchronous AHS: at +200\,ms on Easy, SR is 42.75\% versus 44.75\%.

\FloatBarrier
\section{Conclusion}

ChunkTrust adapts robot-policy execution horizons using action-expert evidence.
AHS combines spectral stability, inter-chunk continuity, and online temporal
memory, while QHA learns a context-conditioned horizon prior from the same
evidence without changing the base policy. Experiments support aggregate gains across two
simulation benchmarks and real-world manipulation, while transfer, ablation,
and cost analyses characterize where adaptation helps. The method requires
accessible generation traces, and gains are not uniform across tasks.
The evaluation also separates per-replan overhead from episode-level cost:
a small selector cost does not imply identical total rollout time.
Task breakdowns likewise expose regressions that aggregate gains can conceal,
particularly when a learned prior is combined with online evidence.
Future work includes multi-chunk horizon inference, cross-policy transfer
of the learned prior, and broader real-world validation across hardware
and domain shifts.

\label{migration:main-text-end}

\clearpage
\bibliographystyle{iclr2027_conference}
\bibliography{references}

\newpage
\appendix
\raggedbottom

\begin{center}
  {\Large\bfseries ChunkTrust: Adapting Execution Horizons\\
  for Robot Policies with Action-Expert Evidence}\\[0.5em]
  {\Large\bfseries Appendix}
\end{center}
\vspace{1em}

\section{Experimental Setup}
\label{sec:exp-setup}

We specify evaluation cohorts, checkpoints, scoring criteria, and implementation
settings for the results in Sec.~\ref{sec:experiments}.

\subsection{Real-world Experiments}
\label{sec:real-world-setup}

\begin{figure}[!t]
  \centering
  \includegraphics[width=0.76\textwidth]{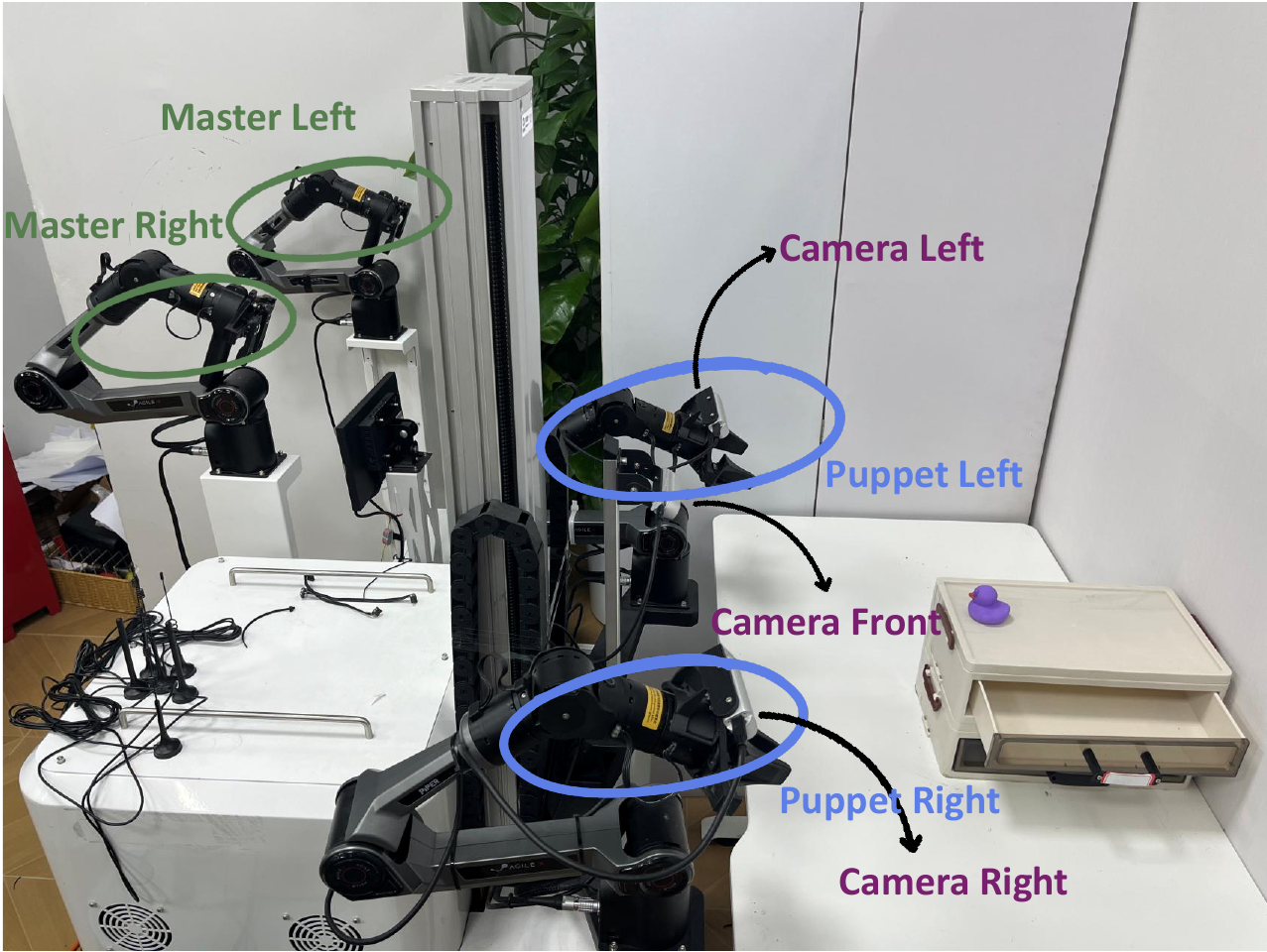}
  \caption{\textbf{Real-world robot platform.}
  AgileX COBOT Magic configured as an ALOHA-style bimanual system for
  teleoperation, data collection, and autonomous policy execution.}
  \label{fig:app-agilex-cobot-magic}
\end{figure}

\paragraph{Hardware Setup.}
We conduct real-world experiments on an AgileX COBOT Magic platform configured
as an ALOHA-style bimanual system~\cite{fu2024mobile,zhaolearning}, as shown in
Fig.~\ref{fig:app-agilex-cobot-magic}.  The platform consists of four 6-DoF Piper
arms, with two leader arms used for human teleoperation and two follower arms
used for data collection and autonomous policy execution.
The perception system includes three RealSense D435 cameras: one front-view
camera and two wrist-mounted cameras, one on each follower arm.

\paragraph{Tasks and evaluation.}
We collect 200 human-teleoperated demonstrations for each of four bimanual
tasks and evaluate $\pi_{0.5}$ with fixed $K=25$ or AHS over 15 rollouts per
task. Data are recorded at 30~FPS, with task-relevant object positions varied
across evaluation rollouts. Instructions are ``Fold the towel with both
arms,'' ``Move the bread to the plate with both arms,'' ``Move the drink to
the basket with both arms,'' and ``Put the duck toy into the drawer with
both arms.'' These tasks cover approach, contact, transport, alignment,
and placement (Fig.~\ref{fig:real-robot-tasks}).

\paragraph{Process scores.}
Each sub-step receives 0 for failure, 0.5 for recovered or imperfect
completion, and 1 for smooth, accurate completion. We average equally
over the task's sub-steps and its rollouts, reporting the result as a
percentage. Table~\ref{tab:real-scoring} lists the task-specific
criteria. Error bars in Fig.~\ref{fig:real-robot-tasks}(b) show
$\pm s/\sqrt{15}$, where $s$ is the sample standard deviation of the
15 normalized rollout scores for each task and method.
In the drawer task, the arm assignment depends on the layout:
one arm opens/closes the drawer and the other manipulates the toy.

\begin{table}[!t]
\centering\small
\caption{\textbf{Real-world scoring criteria.} Scoring is independent of arm assignment.}
\label{tab:real-scoring}
\setlength{\tabcolsep}{4pt}
\renewcommand{\arraystretch}{1.12}
\begin{tabular}{@{}>{\raggedright\arraybackslash}p{0.23\linewidth}>{\raggedright\arraybackslash}p{0.21\linewidth}>{\raggedright\arraybackslash}p{0.26\linewidth}>{\raggedright\arraybackslash}p{0.23\linewidth}@{}}
\toprule
Task and sub-step & 0 & 0.5 & 1\\
\midrule
Bread: grasp & Fails to grasp & Multiple attempts & Smooth first attempt\\
Bread: handover & Handover fails & Unstable/awkward receiving grasp & Stable, aligned grasp\\
Bread: place on plate & Not placed & Poor alignment or rough placement & Clean placement\\
\addlinespace
Drink: push & Only tilts, no useful displacement & Acceptable position, tilt/misalignment & Good position, stable alignment\\
Drink: grasp & Fails to grasp & Multiple attempts & Smooth first attempt\\
Drink: place in basket & Fails or drops drink & Rough placement/collision & Clean placement\\
\addlinespace
Towel: first fold, second fold & Not folded over & Folded, misaligned & Folded, well aligned\\
\addlinespace
Duck: open drawer, grasp toy, place toy, close drawer & Sub-step fails & Multiple attempts & Smooth first attempt\\
\bottomrule
\end{tabular}
\end{table}

\subsection{Simulation Experiments}

\textbf{RoboTwin2.0}~\cite{chen2025robotwin} contains 50 bimanual manipulation
tasks with strong domain randomization.  The complete-suite evaluation in
Tab.~\ref{tab:ahs-summary} uses one multitask $\pi_{0.5}$ checkpoint on all
50 tasks, with 20 rollouts per task, setting, and method (2,000 per method).
The eight-task evaluation uses 100 rollouts per task, setting, and method
(1,600 per method), on Blocks Ranking RGB, Handover Block,
Handover Mic, Hanging Mug, Place A2B Left, Place Bread Basket, Place Bread
Skillet, and Place Can Basket.  Each task is evaluated under two settings:
\textit{Easy} (clean scenes) and \textit{Hard}
(randomized object poses, lighting, and distractor placement).  Results report
success rate, averaged equally across the tasks and settings in each cohort.

\textbf{RoboCasa GR1 Tabletop}~\cite{nasiriany2024robocasa} provides 24 pick
and-place tasks spanning everyday object categories and novel source--target
combinations. We evaluate all 24 tasks, using 50 rollouts per task for the
$\pi_{0.5}$ control. Results for the other policies use their respective
evaluation budgets and reported numerical precision. Task
success is determined by the native RoboCasa success checker.

\subsection{Base Policy Checkpoints}

For the eight-task RoboTwin2.0 evaluation, checkpoints follow the protocol
of each base policy:

\begin{itemize}
  \item $\pi_0$~\cite{black2024pi_0}: we train LoRA adapters~\cite{hu2022lora}
    with the released RoboTwin2.0 fine-tuning protocol.  For each evaluated
    task, one adapter is trained on that task's clean50 demonstrations and then
    evaluated.  The action chunk size is $H=50$, with a flow-matching action
    expert~\cite{lipmanflow} using 10 sampling steps.
  \item $\pi_{0.5}$~\cite{intelligence2025pi_}: we use the official RoboTwin2.0
    fine-tuning recipe for full-parameter fine-tuning.  For each evaluated task,
    one task-specific checkpoint is trained on clean50 demonstrations and then
    evaluated.  The action chunk size is $H=50$, with a flow-matching action
    expert~\cite{lipmanflow}.
  \item X-VLA~\cite{zheng2026xvla}: we evaluate the released X-VLA RoboTwin2.0
    checkpoint.  The action chunk size is $H=30$, with a flow-matching action
    expert.
  \item Fast-WAM~\cite{yuan2026fast}: we evaluate the released Fast-WAM
    multitask RoboTwin2.0 checkpoint on our eight-task subset, rather than
    post-training a separate policy for each task.  The action chunk size
    is $H=32$, with a flow-matching action expert.
\end{itemize}

\paragraph{Multitask $\pi_{0.5}$ checkpoints.}
The complete-suite controls initialize from the official
\texttt{pi05\_base} and use benchmark-specific full-parameter post-training.
RoboTwin2.0 uses 50 clean demonstrations per task (2,500 trajectories and
549,787 frames). RoboCasa GR1 uses 24,000 demonstrations and 6,020,058
frames across 24 tasks.  Both use task-uniform sampling, one global
quantile normalizer per benchmark, global batch size 256, seed 42, and
eight H100 GPUs for training, with the checkpoint fixed at step 30,000.
RoboTwin uses $H=50$ and Base $K=50$, while RoboCasa uses $H=16$ and Base $K=16$.
The latter maps the raw 44-dimensional GR1 interface to 29 effective
absolute controls in the padded policy interface.  These controls share
a policy architecture, not weights across benchmarks. Within each
benchmark, Base and AHS use identical policy weights.  Their evaluation
runs on one RTX 4090.

For the other RoboCasa policies, Isaac-GR00T~N1.5 and
Isaac-GR00T~N1.6~\cite{bjorck2025gr00t} use released base checkpoints
without additional benchmark post-training.  Qwen3GR00T uses the released
24-task multitask-post-trained GR1 checkpoint from
StarVLA~\cite{ye2026starvla,community2026starvla}.  Thus, ``zero-shot'' in
Tab.~\ref{tab:ahs-summary} describes benchmark adaptation, not an absence
of robot pretraining.

\subsection{AHS Hyperparameter Configurations}

Candidate sets and continuity windows depend on the policy and benchmark:
\begin{center}
\small
\begin{tabular}{@{}lcc@{}}
\toprule
Evaluation & Candidate horizons $\mathcal K$ & Continuity window $W_h$\\
\midrule
RoboTwin2.0, $\pi_0$ and $\pi_{0.5}$ & $\{10,20,30,40,50\}$ & 60\\
RoboTwin2.0, X-VLA and Fast-WAM & $\{10,20,30\}$ & 40\\
RoboCasa GR1, all policies & $\{4,8,12,16\}$ & 20\\
\bottomrule
\end{tabular}
\end{center}
The scaling ablation compares $\mathcal K_5=\{10,20,30,40,50\}$,
$\mathcal K_{10}=\{5,10,\ldots,50\}$, and $\mathcal K_{50}=\{1,\ldots,50\}$.
Shared defaults are $\alpha_{\mathrm{intra}}=\beta_{\mathrm{inter}}=4$,
$\lambda=0.5$, $a_0=b_0=1$, $\rho_f=0.99$, $\eta=1$,
and $\sigma_K=10$. The eight-task study uses expected-round
selection at $T_{\mathrm{sel}}=0.8$, with Thompson exploration probability
$\epsilon_{\mathrm{exp}}=0.05$. The 50-task evaluation uses
$T_{\mathrm{sel}}=1.0$, and the RoboCasa $\pi_{0.5}$ control uses 0.8.

Spectral scoring zero-pads each velocity prefix to $H$ before FFT, uses
the first $W_\tau=3$ denoising steps as its reference, window RMS for
$z$, and cutoff fraction $c_{\mathrm{cut}}=0.25$.
Without sufficient executed history, AHS uses intra-chunk-only scoring
($\lambda=0$). Costs appear in Appendix~\ref{sec:runtime-costs}.

\subsection{Evaluation Protocol}

For every (base policy, task, setting) combination, we run a fixed number of
rollouts with AHS enabled.  Each rollout starts from
the standard initial state distribution of the benchmark.  The online AHS posterior is
reset to its prior ($a_0=b_0=1$) at the beginning of each episode, so its
rollout history is episode-local. When QHA is used, its learned weights
are reused across episodes without online training.  Success is determined by the
benchmark's native success checker.  We report the raw success rate (percentage
of successful rollouts) and equal-weight averages over the stated task
and setting groups.  Absolute differences are in percentage points.
The $\Delta$ (pp) columns subtract the corresponding success-rate
percentages, rather than reporting relative percentage changes.

In the 50-task evaluation, 62 of the 100 task--setting conditions use
exact same-seed and instruction pairing between Base and AHS.  The other
38 use deterministic, method-specific fallback identities selected
without outcome information after repeated scene-construction failures.
The full-suite averages include both groups. Identical episode identities
are not assumed for the fallback group.

For the main RoboTwin2.0 benchmarks, Fast-WAM inference uses one NVIDIA H100
(80~GB); $\pi_0$, $\pi_{0.5}$, and X-VLA use one RTX 4090 (24~GB).
The RoboCasa $\pi_{0.5}$ control also uses one RTX 4090. Hardware details for
separate runtime profiles are discussed in Appendix~\ref{sec:runtime-costs}.
Replanning frequency depends on the selected horizon and execution scheduler.

\begin{figure}[!t]
  \centering
  \includegraphics[width=\textwidth]{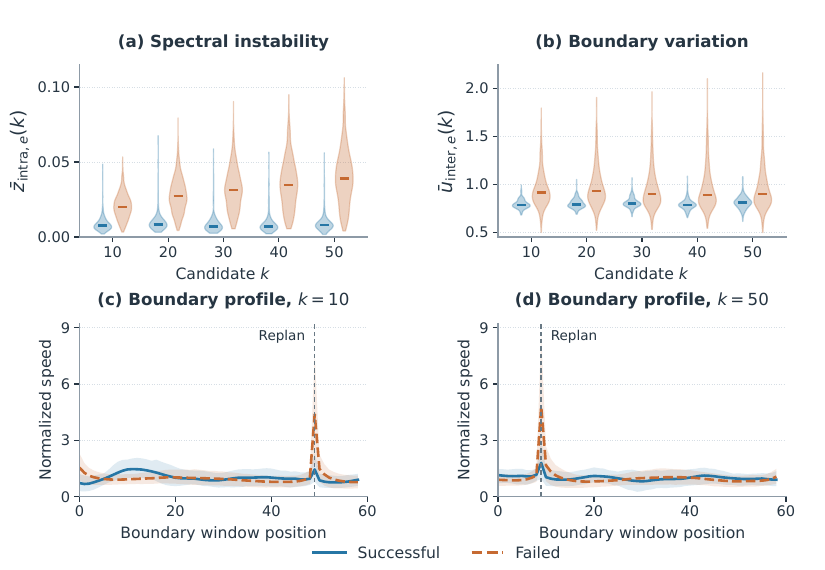}
  \caption{\textbf{Full evidence distributions and boundary behavior.}
  \textbf{(a,b)} Episode means of the operational scores, with violin marks indicating
  medians. Widths show within-group density, not relative sample counts.
  \textbf{(c,d)} Normalized speed profiles for candidate $k=10$ and $k=50$,
  with dashed vertical lines marking the history--prediction join. Curves show
  means and bands show P10--P90 across episode profiles, not confidence
  intervals or the IQR bands used in the main figure.}
  \label{fig:metric-rationality-detail}
\end{figure}

\subsection{QHA Training Configurations}
\label{sec:qha-details}

\paragraph{Training-task protocol (Tab.~\ref{tab:qha-transfer}A).}
We train one shared QHA per policy family
($\pi_0$ or $\pi_{0.5}$) across all eight tasks, keeping the task-specific
base generators frozen. Training uses their clean50 Aloha-AgileX LeRobot
datasets, with task-balanced batches of 256 (32 per task). Inputs comprise
three RGB streams, robot state, action windows, VLM context tokens
$\mathbf C_t$ with mask $\mathbf M_t$, and action-latent tokens
$\mathbf Z_t^A$ from the final chunk. The action-query window includes
64 history and 50 future steps. The frozen checkpoint generates the
chunk $\hat{\mathbf A}_t$, evidence $\mathcal F_t$, and dense teacher
$\mu_t^\star$ online. The teacher follows Eq.~\ref{eq:teacher-distribution}
over $h=1,\ldots,50$ at $T_{\mathrm{teach}}=1$, using current evidence
without episode-local Beta memory. The six-task held-out protocol is
specified separately in Appendix~\ref{sec:qha-heldout}.

\paragraph{Architecture and optimization.}
Context and action tokens are projected into a shared $d=256$ space
with positional encodings. One bridge layer updates eight learned queries
by cross-attending to masked context, then action tokens, followed by
self-attention, each with residual connections~\cite{vaswani2017attention}.
It uses eight attention heads, zero dropout, and an enabled fusion gate.
Query pooling and an MLP produce 50 logits followed by a softmax.
We minimize $\operatorname{KL}(\mu_t^\star\Vert p_\phi)$ with weight 1
and numerical floor $10^{-6}$. Only QHA parameters are updated and the
base flow-matching loss is zero. AdamW runs for 10,000 steps with batch
size 256, global-norm clipping 1, weight decay $10^{-10}$, and EMA 0.99.
The warmup-cosine schedule uses 1,000 warmup steps, peak learning rate
$5\times10^{-5}$, and final rate $5\times10^{-6}$.

\paragraph{Deployment (Tab.~\ref{tab:qha-transfer}A).}
Each policy family uses its QHA checkpoint saved at step 10,000 after joint
training on the eight tasks, with prior strength $\gamma=1$ in Eq.~\ref{eq:qha-prior-main}.
Online evidence $q_{\mathrm{mix},t}(k)$ is computed on
$\mathcal K=\{10,20,30,40,50\}$ and interpolated onto
$\{1,\ldots,H\}$ before Beta sampling. Both this sparse-evidence variant
and the dense-evidence held-out variant maintain dense Beta states and
apply temperature scaling, expected-round selection, and uniform exploration
as in Sec.~\ref{subsec:test-time-method}.
Posterior feedback uses prior-weighted quality
$\tilde q_{\mathrm{mix},t}(h)p_\phi(h)^\gamma$.

\paragraph{Held-out deployment (Tab.~\ref{tab:qha-transfer}B).}
This evaluation instead computes dense evidence over $h=1,\ldots,50$,
with $\gamma=1$ and expected-round selection at $T_{\mathrm{sel}}=1$.
Appendix~\ref{sec:qha-heldout} specifies the six-task training split
and checkpoints.

\begin{figure}[!t]
  \centering
  \includegraphics[width=\textwidth]{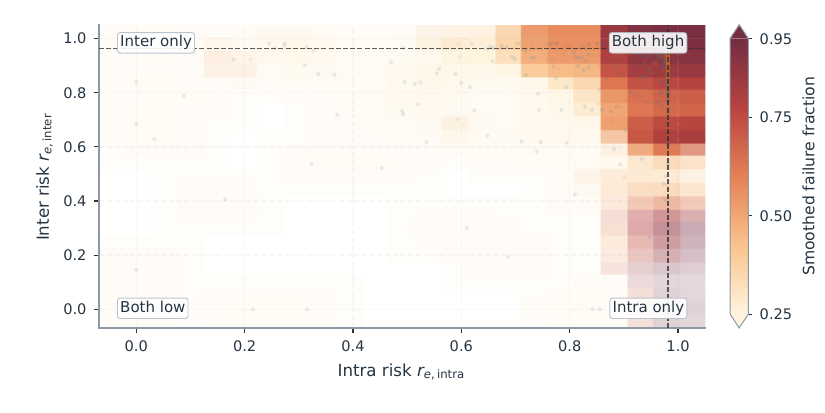}
  \caption{\textbf{Joint-risk landscape on the same 1,600 episodes.}
  Dashed lines mark the original median thresholds. Points denote episodes,
  with outcome colors as in Fig.~\ref{fig:metric-rationality-detail}. Color shows the
  smoothed local failure fraction. Lower opacity denotes lower smoothed
  occupancy. The color scale saturates below 0.25 and above 0.95.
  The map is descriptive. Group failure rates below are computed directly
  from episode outcomes, not read from the smoothed colors.}
  \label{fig:app-joint-risk-map}
\end{figure}

\section{Action-Expert Evidence Diagnostics}
\label{app:fig2-diagnostics}

\subsection{Evidence Distributions and Boundary Profiles}
\label{app:evidence-distributions}

\paragraph{Data and aggregation.}
We analyze the same 1,600 $\pi_0$ RoboTwin2.0 episodes as
Fig.~\ref{fig:metric-rationality-fixedk50}: eight tasks, two settings,
and 100 episodes per task--setting pair, with 291 successes and 1,309 failures.
All episodes execute fixed $K=H=50$.
The candidates $k\in\{10,20,30,40,50\}$ are scored on these recorded
traces. They are not five separate executed-horizon experiments.

For each candidate, we average finite scores over replans within an episode,
then report medians and IQRs over episode means. The valid counts can differ
between the two metrics because inter-chunk evidence requires executed
history. The intra score follows Eq.~\ref{eq:intra-raw}, with prefixes
zero-padded to $H$, a three-step baseline, and RMS over all ten denoising
steps. Recomputed scores agree with the recorded values within $1.5\times10^{-8}$.

For boundary profiles, we compute first-difference speeds in the
$W_h=60$ stitched window of Eq.~\ref{eq:stitched-window}, normalize by
the window mean, and average within each episode before summarizing
across episodes. This prevents longer episodes receiving extra weight.

\subsection{Joint Risk and Episode Outcomes}
\label{app:joint-risk}

At the executed $K=50$, each episode's intra/inter risk is the 75th percentile
of $1-q_{\mathrm{intra},t}(50)$ or $1-q_{\mathrm{inter},t}(50)$ over valid
replans. Pooled median thresholds are 0.98168436 and 0.96278395, respectively.
Values strictly below the threshold are low and ties are high, so group
sizes can differ.

\begin{table}[!t]
  \centering
  \caption{\textbf{Exact median-split group outcomes.}
  Counts and failure rates reproduce Fig.~\ref{fig:metric-rationality-fixedk50}(c).}
  \label{tab:app-risk-groups}
  \small
  \begin{tabular}{lccrrr}
    \toprule
    Group & Intra risk & Inter risk & Episodes & Failures & Failure rate\\
    \midrule
    Both low   & Low  & Low  & 313 & 157 & 50.2\%\\
    Inter only & Low  & High & 193 & 144 & 74.6\%\\
    Intra only & High & Low  & 487 & 417 & 85.6\%\\
    Both high  & High & High & 607 & 591 & 97.4\%\\
    \bottomrule
  \end{tabular}
\end{table}

The map uses a $23\times23$ grid with 7\% range padding and separable
kernel $[1,4,6,4,1]/16$. Success/failure counts are smoothed separately
before division (stabilizer $10^{-12}$). Opacity scales to the 90th
occupancy percentile, hiding cells below $10^{-3}$.

\paragraph{Scope of the evidence.}
These are pooled, retrospective associations from one policy and a fixed
execution horizon. They do not establish causality, calibrated failure
prediction, within-task effects independent of task difficulty, or how long a
particular prefix remains reliable at the current replan.
The benefit of adaptive execution is evaluated separately in the policy
comparisons and ablations, not inferred from this heatmap.

\section{Full Simulation Benchmark Results}
\label{sec:full-simulation-results}

\subsection{QHA Transfer to Held-out Tasks}
\label{sec:qha-heldout}

For Tab.~\ref{tab:qha-transfer}B, QHA is trained on Handover Block,
Handover Mic, Hanging Mug, Place A2B Left, Place Bread Skillet, and Place
Can Basket. Blocks Ranking RGB and Place Bread Basket are held out.
The QHA head uses step 10,000, and frozen task-specific $\pi_{0.5}$ bases
use step-20,000 clean50 checkpoints without quantile normalization.
All selectors use $\{1,\ldots,50\}$ and expected-round selection at
temperature 1.0. AHS+QHA uses $\gamma=1.0$. This cohort differs from
the eight-task study in Tab.~\ref{tab:qha-transfer}A.

AHS, QHA-only, and AHS+QHA share 100 episode identities per task and
setting (400 per method). The Base results in Tab.~\ref{tab:qha-transfer}B
come from a separate evaluation outside this paired cohort. The transfer comparison is AHS+QHA versus
AHS. Table~\ref{tab:heldout-full} retains all four conditions: fusion
improves two and reduces success in two. The aggregate gain does not
establish uniform improvement or transfer of the base policy.

\begin{table}[!t]
\centering
\caption{\textbf{Complete held-out comparison.}
Success rate (\%) for the $\pi_{0.5}$ policy with QHA trained on six tasks and evaluated
on two held-out tasks. Each task and setting uses 100 paired episodes.
$\Delta=\mathrm{SR}_{\mathrm{AHS+QHA}}-\mathrm{SR}_{\mathrm{AHS}}$.}
\label{tab:heldout-full}
\begingroup
\fontsize{9}{10.8}\selectfont
\setlength{\tabcolsep}{4pt}
\renewcommand{\arraystretch}{1.05}
\begin{tabular*}{\linewidth}{@{\extracolsep{\fill}}ll>{\columncolor{cttableahs}}rr>{\columncolor{cttableqha}}rr@{}}
\toprule
Task & Setting & AHS & QHA-only & AHS+QHA & $\Delta$ (pp)\\
\midrule
Blocks Ranking RGB & Easy & 44.00 & 48.00 & \textbf{57.00} & \ctgain{+13.00} \\
Blocks Ranking RGB & Hard & \textbf{33.00} & 29.00 & 30.00 & \ctloss{-3.00} \\
Place Bread Basket & Easy & \textbf{52.00} & 50.00 & 49.00 & \ctloss{-3.00} \\
Place Bread Basket & Hard & 31.00 & 31.00 & \textbf{35.00} & \ctgain{+4.00} \\
\midrule
\textbf{Overall} & Both & 40.00 & 39.50 & \textbf{42.75} & \ctgain{+2.75} \\
\bottomrule
\end{tabular*}\par
\endgroup
\end{table}

\paragraph{Same-state decision disagreement.}
In AHS+QHA traces, the QHA and AHS component choices differ at 77.02\%
of 12,573 replans, with a mean absolute gap of 1.83 action steps.
Easy contributes 5,533 replans (77.17\%, 1.84 steps) and Hard 7,040
(76.90\%, 1.82 steps). This measures differences at the same states,
without establishing complementarity or explaining success gains.

\subsection{Complete 50-task RoboTwin2.0 Evaluation}
\label{sec:robotwin50-results}

\begin{table}[!t]
\centering
\caption{\textbf{Complete 50-task RoboTwin2.0 results.}
Success rate (\%) for the shared multitask $\pi_{0.5}$ checkpoint, with
20 episodes per task, setting, and method. $\Delta$ is AHS minus Base,
averaged over Easy and Hard.}\label{tab:robotwin50-full}
\fontsize{9}{10.8}\selectfont
\renewcommand{\arraystretch}{0.95}
\setlength{\tabcolsep}{4pt}
\begin{tabular}{@{}lrrrrr@{}}
\toprule
\multicolumn{1}{l}{Task} & \multicolumn{2}{c}{Easy} & \multicolumn{2}{c}{Hard} & $\Delta$ (pp)\\
\cmidrule(lr){2-3}\cmidrule(lr){4-5}
& Base & AHS & Base & AHS & \\
\midrule
adjust bottle & 100.00 & 100.00 & 90.00 & 95.00 & \ctgain{+2.50} \\
beat block hammer & 70.00 & 80.00 & 15.00 & 50.00 & \ctgain{+22.50} \\
blocks ranking rgb & 80.00 & 95.00 & 40.00 & 70.00 & \ctgain{+22.50} \\
blocks ranking size & 25.00 & 45.00 & 25.00 & 35.00 & \ctgain{+15.00} \\
click alarmclock & 60.00 & 65.00 & 45.00 & 45.00 & \ctgain{+2.50} \\
click bell & 35.00 & 70.00 & 55.00 & 55.00 & \ctgain{+17.50} \\
dump bin bigbin & 95.00 & 95.00 & 80.00 & 75.00 & \ctloss{-2.50} \\
grab roller & 100.00 & 100.00 & 90.00 & 90.00 & \ctgain{+0.00} \\
handover block & 65.00 & 65.00 & 10.00 & 35.00 & \ctgain{+12.50} \\
handover mic & 100.00 & 100.00 & 15.00 & 15.00 & \ctgain{+0.00} \\
hanging mug & 20.00 & 30.00 & 5.00 & 10.00 & \ctgain{+7.50} \\
lift pot & 65.00 & 70.00 & 30.00 & 35.00 & \ctgain{+5.00} \\
move can pot & 50.00 & 75.00 & 10.00 & 25.00 & \ctgain{+20.00} \\
move pillbottle pad & 55.00 & 80.00 & 30.00 & 30.00 & \ctgain{+12.50} \\
move playingcard away & 90.00 & 80.00 & 75.00 & 75.00 & \ctloss{-5.00} \\
move stapler pad & 20.00 & 35.00 & 20.00 & 5.00 & \ctgain{+0.00} \\
open laptop & 95.00 & 90.00 & 85.00 & 70.00 & \ctloss{-10.00} \\
open microwave & 85.00 & 50.00 & 35.00 & 25.00 & \ctloss{-22.50} \\
pick diverse bottles & 45.00 & 55.00 & 35.00 & 70.00 & \ctgain{+22.50} \\
pick dual bottles & 55.00 & 70.00 & 75.00 & 75.00 & \ctgain{+7.50} \\
place a2b left & 75.00 & 70.00 & 55.00 & 65.00 & \ctgain{+2.50} \\
place a2b right & 70.00 & 75.00 & 50.00 & 50.00 & \ctgain{+2.50} \\
place bread basket & 55.00 & 75.00 & 55.00 & 55.00 & \ctgain{+10.00} \\
place bread skillet & 35.00 & 55.00 & 60.00 & 55.00 & \ctgain{+7.50} \\
place burger fries & 95.00 & 95.00 & 90.00 & 100.00 & \ctgain{+5.00} \\
place can basket & 40.00 & 85.00 & 0.00 & 35.00 & \ctgain{+40.00} \\
place cans plasticbox & 40.00 & 85.00 & 55.00 & 80.00 & \ctgain{+35.00} \\
place container plate & 75.00 & 90.00 & 80.00 & 80.00 & \ctgain{+7.50} \\
place dual shoes & 75.00 & 85.00 & 50.00 & 60.00 & \ctgain{+10.00} \\
place empty cup & 100.00 & 100.00 & 65.00 & 70.00 & \ctgain{+2.50} \\
place fan & 75.00 & 75.00 & 45.00 & 40.00 & \ctloss{-2.50} \\
place mouse pad & 15.00 & 45.00 & 25.00 & 25.00 & \ctgain{+15.00} \\
place object basket & 50.00 & 60.00 & 20.00 & 35.00 & \ctgain{+12.50} \\
place object scale & 60.00 & 85.00 & 45.00 & 50.00 & \ctgain{+15.00} \\
place object stand & 85.00 & 80.00 & 55.00 & 90.00 & \ctgain{+15.00} \\
place phone stand & 65.00 & 65.00 & 45.00 & 35.00 & \ctloss{-5.00} \\
place shoe & 85.00 & 90.00 & 75.00 & 55.00 & \ctloss{-7.50} \\
press stapler & 100.00 & 90.00 & 60.00 & 70.00 & \ctgain{+0.00} \\
put bottles dustbin & 45.00 & 65.00 & 50.00 & 55.00 & \ctgain{+12.50} \\
put object cabinet & 25.00 & 35.00 & 20.00 & 15.00 & \ctgain{+2.50} \\
rotate qrcode & 80.00 & 85.00 & 30.00 & 30.00 & \ctgain{+2.50} \\
scan object & 20.00 & 40.00 & 20.00 & 50.00 & \ctgain{+25.00} \\
shake bottle & 100.00 & 100.00 & 100.00 & 100.00 & \ctgain{+0.00} \\
shake bottle horizontally & 100.00 & 100.00 & 100.00 & 100.00 & \ctgain{+0.00} \\
stack blocks three & 75.00 & 55.00 & 35.00 & 45.00 & \ctloss{-5.00} \\
stack blocks two & 85.00 & 100.00 & 80.00 & 90.00 & \ctgain{+12.50} \\
stack bowls three & 60.00 & 65.00 & 35.00 & 30.00 & \ctgain{+0.00} \\
stack bowls two & 100.00 & 85.00 & 85.00 & 90.00 & \ctloss{-5.00} \\
stamp seal & 35.00 & 30.00 & 25.00 & 30.00 & \ctgain{+0.00} \\
turn switch & 40.00 & 40.00 & 25.00 & 25.00 & \ctgain{+0.00} \\
\midrule
\textbf{Mean by setting} & 65.40 & 73.10 & 48.00 & 53.90 & \ctgain{+6.80}\\
\textbf{Overall} & \multicolumn{2}{c}{Base: 56.70} & \multicolumn{2}{c}{AHS: 63.50} & \ctgain{+6.80}\\
\bottomrule
\end{tabular}
\end{table}

Table~\ref{tab:robotwin50-full} expands the multitask $\pi_{0.5}$ row of
Tab.~\ref{tab:ahs-summary}.  Base and AHS succeed in 1,134 and 1,270 of
2,000 episodes, respectively, giving 56.70\% and 63.50\% overall.
Easy success increases from 65.40\% to 73.10\%, and Hard from 48.00\% to
53.90\%.  All 50 tasks are retained, including nine tasks with a negative
change after averaging the two settings.

\paragraph{Statistical robustness and pairing scope.}
The full-matrix gain is 6.80 percentage points, with a task-cluster
95\% bootstrap interval of $[3.75,\,9.95]$. This interval resamples
the 50 tasks, retaining both settings and methods within each task.
Only 62 of 100 task--setting cells preserve exact seed-and-instruction
pairing, while the remaining 38 cells use deterministic, outcome-blind,
method-specific fallback identities after scene-construction failures.
The 1,240 exact episode pairs yield a gain of 7.58 percentage points
with a paired 95\% interval of $[5.08,\,10.08]$, resampling episodes
within these cells. Both intervals use 10,000 bootstrap draws, but their
statistical units and populations differ: the paired interval applies
only to the exact-identity subset, not the full matrix.

\subsection{Eight-task Policy-family Evaluation}

The task-specific $\pi_0$ and $\pi_{0.5}$ matrices appear in
Tab.~\ref{tab:qha-transfer}A.  Table~\ref{tab:robotwin-additional} provides
the corresponding Base--AHS breakdown for Fast-WAM and
X-VLA~\cite{zheng2026xvla}, evaluated on the same eight tasks with
100 rollouts per task and setting.  Both use $\mathcal K=\{10,20,30\}$.
Each Base--AHS comparison uses the same released policy checkpoint.
Fast-WAM's overall success increases from 86.81\% to 88.13\%.
For X-VLA, AHS slightly improves overall success from 47.44\% to 47.75\%:
Easy increases from 74.50\% to 75.75\%, while Hard decreases from
20.38\% to 19.75\%.  Individual task regressions are retained for both
policies.

\begin{table}[!t]
\centering
\caption{\textbf{Fast-WAM and X-VLA per-task success rates on RoboTwin2.0.}
Success rate (\%) over 100 rollouts per task and setting. Easy and Hard
denote clean and randomized settings. Each Base--AHS comparison fixes
the policy checkpoint. Yellow columns use AHS, and bold marks the better
method within each policy and setting, including ties.}
\label{tab:robotwin-additional}
\begingroup
\fontsize{9}{10.8}\selectfont
\setlength{\tabcolsep}{2.5pt}
\renewcommand{\arraystretch}{1.05}
\begin{tabular*}{\linewidth}{@{\extracolsep{\fill}}lrr>{\columncolor{cttableahs}}r>{\columncolor{cttableahs}}rrr>{\columncolor{cttableahs}}r>{\columncolor{cttableahs}}r@{}}
\toprule
& \multicolumn{4}{c}{Fast-WAM~{\footnotesize\citep{yuan2026fast}}} & \multicolumn{4}{c}{X-VLA~{\footnotesize\citep{zheng2026xvla}}}\\
\cmidrule(lr){2-5}\cmidrule(lr){6-9}
Task & \multicolumn{2}{c}{Base} & \multicolumn{2}{c}{AHS}
     & \multicolumn{2}{c}{Base} & \multicolumn{2}{c}{AHS}\\
\cmidrule(lr){2-3}\cmidrule(lr){4-5}\cmidrule(lr){6-7}\cmidrule(lr){8-9}
& Easy & Hard & Easy & Hard & Easy & Hard & Easy & Hard\\
\midrule
Blocks Ranking RGB & 99.00 & 98.00 & \textbf{100.00} & \textbf{100.00} & 87.00 & 34.00 & \textbf{95.00} & \textbf{38.00}\\
Handover Block & \textbf{94.00} & 81.00 & 91.00 & \textbf{82.00} & \textbf{89.00} & \textbf{2.00} & 88.00 & 1.00\\
Handover Mic & \textbf{100.00} & 99.00 & \textbf{100.00} & \textbf{100.00} & 97.00 & \textbf{1.00} & \textbf{100.00} & \textbf{1.00}\\
Hanging Mug & 66.00 & 64.00 & \textbf{71.00} & \textbf{69.00} & 32.00 & \textbf{8.00} & \textbf{35.00} & 4.00\\
Place A2B Left & \textbf{95.00} & \textbf{95.00} & 93.00 & 93.00 & 31.00 & \textbf{25.00} & \textbf{40.00} & 21.00\\
Place Bread Basket & 91.00 & 91.00 & \textbf{93.00} & \textbf{94.00} & \textbf{87.00} & 41.00 & 80.00 & \textbf{42.00}\\
Place Bread Skillet & 91.00 & 91.00 & \textbf{94.00} & \textbf{94.00} & 86.00 & 22.00 & \textbf{88.00} & \textbf{23.00}\\
Place Can Basket & \textbf{71.00} & 63.00 & 69.00 & \textbf{67.00} & \textbf{87.00} & \textbf{30.00} & 80.00 & 28.00\\
\midrule
Mean by setting & 88.38 & 85.25 & \textbf{88.88} & \textbf{87.38} & 74.50 & \textbf{20.38} & \textbf{75.75} & 19.75\\
\textbf{Overall} & \multicolumn{2}{c}{86.81} & \multicolumn{2}{>{\columncolor{cttableahs}}c}{\textbf{88.13}}
& \multicolumn{2}{c}{47.44} & \multicolumn{2}{>{\columncolor{cttableahs}}c}{\textbf{47.75}}\\
\bottomrule
\end{tabular*}\par
\endgroup
\end{table}

\subsection{RoboCasa GR1 Tabletop}

Table~\ref{tab:robocasa-full} reports the full 24-task RoboCasa GR1 Tabletop
breakdown for $\pi_{0.5}$ and the three GR00T-family policies summarized in
Tab.~\ref{tab:ahs-summary}.  QwenFAST (discrete tokens) and QwenPI
(flow-matching action expert) are additional StarVLA
baselines~\cite{ye2026starvla,community2026starvla}, both using
Qwen3VL~\cite{bai2025qwen3}. Neither has an AHS counterpart in this table.
For $\pi_{0.5}$, the 24-task means are 40.08\% for Base and 42.50\% for AHS,
matching Tab.~\ref{tab:ahs-summary}.

\begin{table}[!t]
  \centering
  \small
  \caption{\textbf{Full per-task success rates on RoboCasa GR1 Tabletop.}
  Full 24-task breakdown, with 50 rollouts per task for $\pi_{0.5}$.
  Yellow columns use AHS and parentheses report percentage-point deltas
  over Base. Bold marks the best result in each row, including ties.}
  \label{tab:robocasa-full}
  \setlength{\tabcolsep}{2pt}
  \resizebox{\textwidth}{!}{%
  \begin{tabular}{@{}lccc>{\columncolor{ahsyellow}}cc>{\columncolor{ahsyellow}}cc>{\columncolor{ahsyellow}}cc>{\columncolor{ahsyellow}}c@{}}
    \toprule
    \textbf{Task} & \textbf{\shortstack{QwenFAST\\+Qwen3VL}} & \textbf{\shortstack{QwenPI\\+Qwen3VL}}
    & \multicolumn{2}{c}{\textbf{\shortstack{Isaac-GR00T\\N1.5}}}
    & \multicolumn{2}{c}{\textbf{\shortstack{Isaac-GR00T\\N1.6}}}
    & \multicolumn{2}{c}{\textbf{\shortstack{Qwen3GR00T\\+Qwen3VL}}}
    & \multicolumn{2}{c}{$\boldsymbol{\pi_{0.5}}$} \\
    \cmidrule(lr){4-5}\cmidrule(lr){6-7}\cmidrule(lr){8-9}\cmidrule(lr){10-11}
    & & & Base & AHS & Base & AHS & Base & AHS & Base & AHS \\
    \midrule
    PnPBottleToCabinetClose & 38.0 & 26.0 & 64.0 & \ahscell{\textbf{68.0} \posdelta{+4.0}} & 51.5 & \ahscell{54.0 \posdelta{+2.5}} & 46.0 & \ahscell{64.0 \posdelta{+18.0}} & 64.00 & \ahscell{\textbf{68.00} \posdelta{+4.00}} \\
    PnPCanToDrawerClose & 44.0 & 62.0 & 18.0 & \ahscell{12.0 \negdelta{-6.0}} & 13.0 & \ahscell{12.0 \negdelta{-1.0}} & \textbf{80.0} & \ahscell{\textbf{80.0} \posdelta{+0.0}} & 58.00 & \ahscell{56.00 \negdelta{-2.00}} \\
    PnPCupToDrawerClose & \textbf{56.0} & 42.0 & 12.0 & \ahscell{4.0 \negdelta{-8.0}} & 8.5 & \ahscell{14.0 \posdelta{+5.5}} & 54.0 & \ahscell{52.0 \negdelta{-2.0}} & 34.00 & \ahscell{40.00 \posdelta{+6.00}} \\
    PnPMilkToMicrowaveClose & 44.0 & \textbf{50.0} & 38.0 & \ahscell{34.0 \negdelta{-4.0}} & 14.0 & \ahscell{20.0 \posdelta{+6.0}} & 48.0 & \ahscell{42.0 \negdelta{-6.0}} & 40.00 & \ahscell{44.00 \posdelta{+4.00}} \\
    PnPPotatoToMicrowaveClose & 14.0 & 42.0 & \textbf{54.0} & \ahscell{36.0 \negdelta{-18.0}} & 41.5 & \ahscell{50.0 \posdelta{+8.5}} & 28.0 & \ahscell{28.0 \posdelta{+0.0}} & 22.00 & \ahscell{30.00 \posdelta{+8.00}} \\
    PnPWineToCabinetClose & 14.0 & 32.0 & 16.0 & \ahscell{20.0 \posdelta{+4.0}} & 16.5 & \ahscell{24.0 \posdelta{+7.5}} & 46.0 & \ahscell{52.0 \posdelta{+6.0}} & 52.00 & \ahscell{\textbf{56.00} \posdelta{+4.00}} \\
    \addlinespace[0.35em]
    PnPNovelFromCuttingboardToBasket & 54.0 & 40.0 & 50.0 & \ahscell{52.0 \posdelta{+2.0}} & 58.0 & \ahscell{54.0 \negdelta{-4.0}} & 48.0 & \ahscell{\textbf{70.0} \posdelta{+22.0}} & 34.00 & \ahscell{34.00 \posdelta{+0.00}} \\
    PnPNovelFromCuttingboardToCardboardbox & 42.0 & 46.0 & 36.0 & \ahscell{34.0 \negdelta{-2.0}} & 46.5 & \ahscell{46.0 \negdelta{-0.5}} & 40.0 & \ahscell{\textbf{54.0} \posdelta{+14.0}} & 32.00 & \ahscell{40.00 \posdelta{+8.00}} \\
    PnPNovelFromCuttingboardToPan & 58.0 & 60.0 & 68.0 & \ahscell{64.0 \negdelta{-4.0}} & 68.5 & \ahscell{\textbf{80.0} \posdelta{+11.5}} & 68.0 & \ahscell{\textbf{80.0} \posdelta{+12.0}} & 54.00 & \ahscell{58.00 \posdelta{+4.00}} \\
    PnPNovelFromCuttingboardToPot & 58.0 & 40.0 & 34.0 & \ahscell{56.0 \posdelta{+22.0}} & 65.0 & \ahscell{64.0 \negdelta{-1.0}} & 52.0 & \ahscell{\textbf{76.0} \posdelta{+24.0}} & 46.00 & \ahscell{40.00 \negdelta{-6.00}} \\
    PnPNovelFromCuttingboardToTieredbasket & 40.0 & 44.0 & 46.0 & \ahscell{32.0 \negdelta{-14.0}} & 46.5 & \ahscell{54.0 \posdelta{+7.5}} & \textbf{56.0} & \ahscell{44.0 \negdelta{-12.0}} & 22.00 & \ahscell{28.00 \posdelta{+6.00}} \\
    \addlinespace[0.35em]
    PnPNovelFromPlacematToBasket & 36.0 & 44.0 & 50.0 & \ahscell{46.0 \negdelta{-4.0}} & \textbf{58.5} & \ahscell{48.0 \negdelta{-10.5}} & 42.0 & \ahscell{54.0 \posdelta{+12.0}} & 42.00 & \ahscell{46.00 \posdelta{+4.00}} \\
    PnPNovelFromPlacematToBowl & 38.0 & 52.0 & 50.0 & \ahscell{62.0 \posdelta{+12.0}} & 57.5 & \ahscell{60.0 \posdelta{+2.5}} & 44.0 & \ahscell{\textbf{66.0} \posdelta{+22.0}} & 34.00 & \ahscell{32.00 \negdelta{-2.00}} \\
    PnPNovelFromPlacematToPlate & 42.0 & 50.0 & 62.0 & \ahscell{66.0 \posdelta{+4.0}} & 63.0 & \ahscell{\textbf{82.0} \posdelta{+19.0}} & 48.0 & \ahscell{72.0 \posdelta{+24.0}} & 46.00 & \ahscell{48.00 \posdelta{+2.00}} \\
    PnPNovelFromPlacematToTieredshelf & 18.0 & 28.0 & 14.0 & \ahscell{26.0 \posdelta{+12.0}} & 28.5 & \ahscell{\textbf{36.0} \posdelta{+7.5}} & 18.0 & \ahscell{20.0 \posdelta{+2.0}} & 28.00 & \ahscell{28.00 \posdelta{+0.00}} \\
    \addlinespace[0.35em]
    PnPNovelFromPlateToBowl & 52.0 & 52.0 & 58.0 & \ahscell{58.0 \posdelta{+0.0}} & 57.0 & \ahscell{58.0 \posdelta{+1.0}} & \textbf{60.0} & \ahscell{\textbf{60.0} \posdelta{+0.0}} & 40.00 & \ahscell{44.00 \posdelta{+4.00}} \\
    PnPNovelFromPlateToCardboardbox & 30.0 & 40.0 & 40.0 & \ahscell{48.0 \posdelta{+8.0}} & 43.5 & \ahscell{\textbf{58.0} \posdelta{+14.5}} & 50.0 & \ahscell{54.0 \posdelta{+4.0}} & 28.00 & \ahscell{30.00 \posdelta{+2.00}} \\
    PnPNovelFromPlateToPan & 48.0 & 36.0 & 44.0 & \ahscell{48.0 \posdelta{+4.0}} & 51.0 & \ahscell{\textbf{68.0} \posdelta{+17.0}} & 54.0 & \ahscell{54.0 \posdelta{+0.0}} & 32.00 & \ahscell{36.00 \posdelta{+4.00}} \\
    PnPNovelFromPlateToPlate & 50.0 & 48.0 & 66.0 & \ahscell{74.0 \posdelta{+8.0}} & 78.7 & \ahscell{\textbf{82.0} \posdelta{+3.3}} & 70.0 & \ahscell{74.0 \posdelta{+4.0}} & 54.00 & \ahscell{54.00 \posdelta{+0.00}} \\
    \addlinespace[0.35em]
    PnPNovelFromTrayToCardboardbox & 28.0 & 34.0 & 44.0 & \ahscell{52.0 \posdelta{+8.0}} & 51.5 & \ahscell{48.0 \negdelta{-3.5}} & 38.0 & \ahscell{\textbf{56.0} \posdelta{+18.0}} & 48.00 & \ahscell{50.00 \posdelta{+2.00}} \\
    PnPNovelFromTrayToPlate & 34.0 & 64.0 & 50.0 & \ahscell{60.0 \posdelta{+10.0}} & \textbf{71.0} & \ahscell{68.0 \negdelta{-3.0}} & 56.0 & \ahscell{62.0 \posdelta{+6.0}} & 40.00 & \ahscell{40.00 \posdelta{+0.00}} \\
    PnPNovelFromTrayToPot & 46.0 & 44.0 & 46.0 & \ahscell{50.0 \posdelta{+4.0}} & 64.5 & \ahscell{64.0 \negdelta{-0.5}} & 50.0 & \ahscell{\textbf{66.0} \posdelta{+16.0}} & 54.00 & \ahscell{54.00 \posdelta{+0.00}} \\
    PnPNovelFromTrayToTieredbasket & 36.0 & 50.0 & 44.0 & \ahscell{38.0 \negdelta{-6.0}} & \textbf{57.0} & \ahscell{56.0 \negdelta{-1.0}} & 36.0 & \ahscell{56.0 \posdelta{+20.0}} & 34.00 & \ahscell{36.00 \posdelta{+2.00}} \\
    PnPNovelFromTrayToTieredshelf & 16.0 & 28.0 & 34.0 & \ahscell{38.0 \posdelta{+4.0}} & 31.5 & \ahscell{34.0 \posdelta{+2.5}} & 16.0 & \ahscell{\textbf{44.0} \posdelta{+28.0}} & 24.00 & \ahscell{28.00 \posdelta{+4.00}} \\
    \midrule
    \textbf{Average} & 39.00 & 43.92 & 43.25 & \ahscell{44.92 \posdelta{+1.67}} & 47.61 & \ahscell{51.42 \posdelta{+3.80}} & 47.83 & \ahscell{\textbf{57.50} \posdelta{+9.67}} & 40.08 & \ahscell{42.50 \posdelta{+2.42}} \\
    \bottomrule
  \end{tabular}%
  }
\end{table}

\subsection{Horizon-selection Comparisons}
\label{sec:horizon-selector-comparisons}

Table~\ref{tab:horizon-selector-comparisons} reports additional controls
for horizon selection, separately from the cross-policy results in
Tab.~\ref{tab:ahs-summary}.

\paragraph{Fixed, action-only, and evidence-update selectors.}
The $\pi_0$ study uses Place A2B Left, Place Bread Basket, Place Bread
Skillet, and Place Can Basket under Easy and Hard settings, with 16
paired episodes per task and setting (128 per selector). Instantaneous is
a separate reference. This cohort differs from the 100-rollout studies in
Tabs.~\ref{tab:pi05-qmix-overhead} and~\ref{tab:pi0-ablation}.
The global fixed horizon $K=20$ is selected retrospectively from
previous evaluations, not from a held-out validation set.
Jerk-min uses an action-only smoothness criterion on
$\mathcal K=\{10,20,30,40,50\}$.

The evidence-update variants share this candidate grid and the same
instantaneous evidence, using expected-round selection with temperature
0.8. Instantaneous uses $q_{\mathrm{mix},t}(k)$, whereas EMA maintains
$m_t(k)=\rho_{\mathrm{EMA}}m_{t-1}(k)+(1-\rho_{\mathrm{EMA}})q_{\mathrm{mix},t}(k)$
with $\rho_{\mathrm{EMA}}=0.99$. Neither comparator uses Beta counts or kernel neighborhood
sharing. Full Beta is the shared AHS reference. Results describe success
and call-count trade-offs without exact compute matching. The paired 95\%
SR-difference interval between AHS and Jerk-min includes zero, so this
compact study does not establish an SR advantage.

\begin{table}[!t]
\centering
\fontsize{10}{11.5}\selectfont
\caption{\textbf{Horizon-selection success rates and computational costs.}
RoboTwin2.0 uses $\pi_0$ on four tasks under Easy and Hard (128 episodes
per selector). Instantaneous is a separate reference without timing measurements.
Inference and wall times are seconds per episode. RoboCasa references are
non-paired. Bold/underline mark best/second-best SR among the displayed
methods within each benchmark.}
\label{tab:horizon-selector-comparisons}\label{tab:runtime-costs}
\setlength{\tabcolsep}{3pt}
\begin{tabular*}{\linewidth}{@{\extracolsep{\fill}}lrrrrr@{}}
\toprule
\multicolumn{6}{@{}l}{\textbf{RoboTwin2.0}}\\
Selector & SR (\%) $\uparrow$ & Calls/ep & Policy infer & Total infer & Wall\\
\midrule
Global fixed ($K=20$) & \underline{14.84} & 25.96 & 2.75 & 2.75 & 54.33\\
Jerk-min & 14.06 & 21.14 & 2.28 & 2.28 & 52.01\\
EMA $q_{\mathrm{mix}}$ & 10.94 & 19.72 & 2.05 & 2.07 & 46.72\\
\rowcolor{cttableahs}
AHS (Full Beta) & \textbf{15.63} & 20.71 & 2.26 & 2.28 & 53.47\\
\midrule
Instantaneous $q_{\mathrm{mix}}$ (reference) & 13.28 & 19.58 & --- & --- & ---\\
\bottomrule
\end{tabular*}
\par\vspace{5pt}
\begin{tabular*}{0.8\linewidth}{@{\extracolsep{\fill}}lrrr@{}}
\toprule
\multicolumn{4}{@{}l}{\textbf{RoboCasa GR1 Tabletop}, Qwen3GR00T, 24 tasks}\\
 & Base & AAC-core & AHS\\
\midrule
SR (\%) $\uparrow$ & 47.83 & \underline{52.58} & \textbf{57.50}\\
\bottomrule
\end{tabular*}
\end{table}

\paragraph{AAC-core on RoboCasa.}
We evaluate a joint-space adapter of AAC~\cite{liang2026adaptive}
on Qwen3GR00T over 24 tasks with
50 rollouts per task. Each replan draws 20 action-head samples under
the same observation and instruction. A prefix-entropy elbow and a
movement guard determine $K\in\{2,\ldots,16\}$, and the first sampled
chunk supplies the executed actions. The adapter operates on the
policy's native 29-dimensional absolute joint targets. It is not an
exact reproduction of the published Cartesian-action implementation.
AAC-core obtains 52.58\% success. Base/AHS values from separate evaluations in
Tab.~\ref{tab:ahs-summary} provide non-paired context only. We do not
report a paired difference or infer compute equivalence from policy
call counts.

\section{Runtime and Latency}
\subsection{Policy-call and Runtime Costs}
\label{sec:runtime-costs}

We group runtime measurements by timing scope. Means include
all episodes, including failures. Policy inference excludes separately
timed horizon selection, but total inference includes it. Episode wall
time includes simulation and within-episode overhead, not physical robot
execution time. These are descriptive profiles, not hardware-matched
cross-policy rankings.

Compact-selector timings are included with their success rates in Tab.~\ref{tab:runtime-costs}.

\begin{table}[!t]
\centering
\caption{\textbf{Policy-inference and episode costs for $\pi_{0.5}$.}
RoboTwin2.0 means include failures. Panels use separate cohorts.
Panel B also differs from the success-rate evaluation in
Tab.~\ref{tab:qha-transfer}B.}
\label{tab:qha-runtime}
\begingroup
\fontsize{10}{11.5}\selectfont
\setlength{\tabcolsep}{4pt}
\renewcommand{\arraystretch}{1.06}
\begin{tabular}{@{}p{\dimexpr.34\linewidth-4pt\relax}>{\raggedleft\arraybackslash}p{\dimexpr.18\linewidth-8pt\relax}>{\raggedleft\arraybackslash}p{\dimexpr.24\linewidth-8pt\relax}>{\raggedleft\arraybackslash}p{\dimexpr.24\linewidth-4pt\relax}@{}}
\toprule
 & Calls/ & Policy inference & Episode wall\\
Method & episode & (s/episode) & time (s/episode)\\
\midrule
\multicolumn{4}{@{}l}{\textbf{A. Full 50-task evaluation}\quad 2,000 episodes per method}\\
\addlinespace[2pt]
Base & 8.00 & 0.97 & 38.19\\
\rowcolor{cttableahs}
AHS & 14.50 & 1.55 & 36.83\\
\midrule
\multicolumn{4}{@{}l}{\textbf{B. QHA timing evaluation}\quad 2 held-out tasks, 400 episodes per method}\\
\addlinespace[2pt]
AHS & 33.77 & 3.29 & 68.60\\
QHA-only & 29.81 & 18.87 & 81.53\\
\rowcolor{cttableqha}
AHS+QHA & 31.43 & 19.88 & 83.32\\
\bottomrule
\end{tabular}
\endgroup
\end{table}

\paragraph{Shared timing scope for $\pi_{0.5}$.}
Table~\ref{tab:qha-runtime} reports policy-only inference time. Total inference
time is unavailable for the 50-task study. QHA selection is outside the policy
timer and is not timed separately. The AHS reference in Panel B has a total
inference time of 3.49~s per episode.
Both QHA variants make fewer calls but have higher policy-inference
and wall times than this reference. These timings and the success rates in
Tab.~\ref{tab:qha-transfer}B come from different cohorts and cannot be combined
to estimate success-normalized efficiency. These costs characterize the evaluated
implementation, not an intrinsic QHA cost.

\paragraph{AAC-core sampling cost.}
On RoboCasa GR1 Tabletop with Qwen3GR00T (24 tasks, 1,200 episodes),
AAC-core averages 180.95 calls, 20.98~s of synchronized total inference,
and 47.56~s of wall time per episode. Each call samples 20 action chunks,
giving 4,342,740 chunks in total. No policy-only timing aggregate is
available. Base/AHS references come from separate evaluations and are not
paired with the AAC-core measurements. AAC-core, the 50-task study,
the four-task selector comparison, and the QHA timing evaluation
were run on an NVIDIA RTX~4090 GPU.

\subsection{Latency and Asynchronous Execution}
\label{sec:rtc-latency-audit}

The policy prediction horizon is $H=50$. Fixed $K=40$ is a target execution
budget. RTC can replace a chunk at the first legal waypoint boundary after
readiness. The AHS candidates are $\{10,20,30,40\}$. This cohort uses no QHA
and is separate from the $K=50$ candidate-scaling experiment.

\paragraph{Timing and aggregation.}
For episode $i$, let $a_{ij}=t^{\mathrm{start}}_{ij}-t^{\mathrm{obs}}_{ij}$
be the age of the observation used to generate executed action $j$.
We compute $\bar a_i=n_i^{-1}\sum_j a_{ij}$ and average $\bar a_i$
equally over episodes within each task, then equally over the four tasks,
separately for Easy and Hard.
Failures are included.

Wait s/ep averages recorded hold time per episode. Wait\,\% is
$100\sum_i W_i/\sum_i T_i$, where $T_i$ includes both action execution and
waiting. Calls/ep includes requests discarded at episode termination.
Infer s/ep reports model computation time per episode, amortized across
active batch requests and excluding selector computation.

\begin{table}[!t]
\centering\small
\caption{\textbf{Latency results by task and in aggregate on RoboTwin2.0 Hard.}
100 paired episodes per task/method/delay, including failures.
Costs and mean observation age use +200\,ms. Fixed uses target $K=40$.
Bold/underline mark best/second-best SR, waiting, and age within each group,
including displayed ties. Calls and inference time are unranked.}
\label{tab:rtc-full-panel}\label{tab:rtc-tasks}
\setlength{\tabcolsep}{3.0pt}
\begin{tabular}{@{}lrrrrrrrr@{}}\toprule
Method & \multicolumn{3}{c}{SR (\%) $\uparrow$} & Wait (\%) $\downarrow$ & Wait s/ep $\downarrow$ & Calls/ep & Infer s/ep & Mean age (s) $\downarrow$ \\
\cmidrule(lr){2-4}
& +0\,ms & +100\,ms & +200\,ms & & & & & \\
\midrule
\multicolumn{9}{l}{\textit{All four tasks}} \\
Sync-Fixed & 19.50 & 19.50 & 19.25 & \underline{3.77} & \underline{4.432} & 12.89 & 5.04 & 5.02 \\
Sync-AHS & \underline{26.25} & \underline{24.25} & \underline{23.75} & 6.80 & 7.893 & 22.95 & 8.59 & \textbf{3.00} \\
RTC-Fixed & 23.00 & 21.50 & 22.00 & \textbf{0.31} & \textbf{0.344} & 13.32 & 6.47 & 4.92 \\
RTC-AHS & \textbf{26.50} & \textbf{25.75} & \textbf{26.75} & \textbf{0.31} & \textbf{0.344} & 24.30 & 10.62 & \underline{3.07} \\
\midrule
\multicolumn{9}{l}{\textit{Place A2B Left}} \\
Sync-Fixed & 15.00 & \underline{21.00} & 16.00 & 3.37 & \underline{3.106} & 9.03 & 3.29 & 5.35 \\
Sync-AHS & 20.00 & 16.00 & \underline{20.00} & 5.92 & 5.790 & 16.83 & 6.21 & \textbf{3.30} \\
RTC-Fixed & \textbf{26.00} & \textbf{22.00} & \textbf{22.00} & \underline{0.41} & \textbf{0.344} & 9.53 & 4.09 & 5.15 \\
RTC-AHS & \underline{22.00} & \underline{21.00} & \textbf{22.00} & \textbf{0.37} & \textbf{0.344} & 17.94 & 7.28 & \underline{3.37} \\
\midrule
\multicolumn{9}{l}{\textit{Place Bread Basket}} \\
Sync-Fixed & 26.00 & 25.00 & 25.00 & 3.20 & \underline{5.246} & 15.25 & 6.12 & 5.77 \\
Sync-AHS & \textbf{38.00} & \textbf{34.00} & \underline{32.00} & 6.19 & 9.409 & 27.36 & 10.14 & \textbf{3.20} \\
RTC-Fixed & 28.00 & 29.00 & 29.00 & \textbf{0.22} & \textbf{0.344} & 15.61 & 7.80 & 5.68 \\
RTC-AHS & \underline{37.00} & \underline{31.00} & \textbf{37.00} & \underline{0.24} & \textbf{0.344} & 28.68 & 12.60 & \underline{3.38} \\
\midrule
\multicolumn{9}{l}{\textit{Place Bread Skillet}} \\
Sync-Fixed & \underline{13.00} & 9.00 & \underline{12.00} & \underline{5.15} & 4.097 & 11.91 & 4.51 & 3.89 \\
Sync-AHS & \underline{13.00} & \underline{13.00} & \underline{12.00} & 8.56 & 6.994 & 20.33 & 7.06 & \underline{2.58} \\
RTC-Fixed & \underline{13.00} & 12.00 & 10.00 & \textbf{0.45} & \underline{0.346} & 12.35 & 5.59 & 3.81 \\
RTC-AHS & \textbf{15.00} & \textbf{16.00} & \textbf{13.00} & \textbf{0.45} & \textbf{0.345} & 21.71 & 8.43 & \textbf{2.54} \\
\midrule
\multicolumn{9}{l}{\textit{Place Can Basket}} \\
Sync-Fixed & 24.00 & 23.00 & 24.00 & \underline{3.90} & \underline{5.280} & 15.35 & 6.27 & 5.08 \\
Sync-AHS & \textbf{34.00} & \underline{34.00} & \underline{31.00} & 7.07 & 9.381 & 27.27 & 10.95 & \textbf{2.93} \\
RTC-Fixed & 25.00 & 23.00 & 27.00 & \textbf{0.27} & \textbf{0.344} & 15.78 & 8.39 & 5.02 \\
RTC-AHS & \underline{32.00} & \textbf{35.00} & \textbf{35.00} & \textbf{0.27} & \textbf{0.344} & 28.86 & 14.19 & \underline{3.01} \\
\bottomrule\end{tabular}\end{table}

\paragraph{Success within a time budget.}
For $T_i^{\mathrm{success}}$, set the recorded completion time for a
successful episode and $+\infty$ for a failed episode. Then
\[
  \mathrm{SR}(t)=\frac{100}{4}\sum_{q=1}^{4}
  \frac{1}{100}\sum_{i\in q}
  \mathbf{1}\{T_i^{\mathrm{success}}\leq t\}.
\]
Here $t$ denotes physical time in seconds. Computed separately for each setting,
$\mathrm{SR}(t)$ measures task completion under a time budget,
with all attempted episodes retained in the denominator.

\paragraph{Easy and Hard outcomes.}
Both settings use the same task checkpoints, execution horizon and AHS candidates;
episodes are paired across methods and delays within each setting.
At +200\,ms, RTC reduces AHS waiting by 94.5\% on Easy and 95.6\% on Hard.
Relative to RTC-Fixed, RTC-AHS achieves higher aggregate SR in all six
setting/delay conditions, with observed gains of 3.50--8.25 percentage points.
At +200\,ms, the paired 95\% interval for this gain is
$[-0.75,8.26]$ points on Easy and $[0.75,8.75]$ on Hard.
Relative to Sync-AHS, RTC-AHS changes final SR by $-2.00$ points on Easy
and $+3.00$ points on Hard at +200\,ms. Thus, reduced waiting does not
imply uniformly higher success. Policy calls and model computation
remain higher than for RTC-Fixed (Tabs.~\ref{tab:rtc-full-panel},
\ref{tab:rtc-easy-panel}, and~\ref{tab:rtc-all-costs}).

\begin{table}[!t]
\centering\small
\caption{\textbf{Latency results by task and in aggregate on RoboTwin2.0 Easy.}
100 paired episodes per task/method/delay, including failures.
Costs and mean observation age use +200\,ms. Fixed uses target $K=40$.
Bold/underline mark best/second-best SR, waiting, and age within each group,
including displayed ties. Calls and inference time are unranked.}
\label{tab:rtc-easy-panel}
\setlength{\tabcolsep}{3.0pt}
\begin{tabular}{@{}lrrrrrrrr@{}}\toprule
Method & \multicolumn{3}{c}{SR (\%) $\uparrow$} & Wait (\%) $\downarrow$ & Wait s/ep $\downarrow$ & Calls/ep & Infer s/ep & Mean age (s) $\downarrow$ \\
\cmidrule(lr){2-4}
& +0\,ms & +100\,ms & +200\,ms & & & & & \\
\midrule
\multicolumn{9}{l}{\textit{All four tasks}} \\
Sync-Fixed & 38.00 & 35.50 & 37.00 & \underline{3.90} & \underline{3.856} & 11.21 & 4.38 & 5.12 \\
Sync-AHS & \textbf{46.00} & \underline{46.25} & \textbf{44.75} & 6.63 & 6.310 & 18.35 & 6.90 & \textbf{3.16} \\
RTC-Fixed & \underline{38.75} & 40.25 & 39.00 & \textbf{0.37} & \textbf{0.344} & 11.50 & 5.57 & 5.00 \\
RTC-AHS & \textbf{46.00} & \textbf{48.50} & \underline{42.75} & \textbf{0.37} & \textbf{0.344} & 20.32 & 8.80 & \underline{3.18} \\
\midrule
\multicolumn{9}{l}{\textit{Place A2B Left}} \\
Sync-Fixed & 49.00 & 45.00 & 50.00 & \underline{3.37} & \underline{2.401} & 6.98 & 2.48 & 5.65 \\
Sync-AHS & \textbf{54.00} & \textbf{58.00} & \underline{56.00} & 5.86 & 4.107 & 11.94 & 4.25 & \textbf{3.41} \\
RTC-Fixed & 50.00 & 51.00 & 50.00 & \textbf{0.51} & \textbf{0.344} & 7.55 & 3.36 & 5.44 \\
RTC-AHS & \underline{52.00} & \underline{57.00} & \textbf{57.00} & \textbf{0.51} & \textbf{0.344} & 12.83 & 5.21 & \underline{3.51} \\
\midrule
\multicolumn{9}{l}{\textit{Place Bread Basket}} \\
Sync-Fixed & 40.00 & 35.00 & 35.00 & 3.41 & \underline{4.548} & 13.22 & 5.29 & 5.51 \\
Sync-AHS & \textbf{47.00} & \underline{47.00} & \textbf{49.00} & 6.10 & 7.317 & 21.27 & 8.19 & \textbf{3.28} \\
RTC-Fixed & 41.00 & 42.00 & \underline{41.00} & \textbf{0.28} & \textbf{0.344} & 13.19 & 6.56 & 5.48 \\
RTC-AHS & \underline{46.00} & \textbf{48.00} & \underline{41.00} & \underline{0.29} & \textbf{0.344} & 24.49 & 10.41 & \underline{3.33} \\
\midrule
\multicolumn{9}{l}{\textit{Place Bread Skillet}} \\
Sync-Fixed & 26.00 & 26.00 & 27.00 & 4.98 & \underline{3.643} & 10.59 & 4.02 & 4.25 \\
Sync-AHS & \underline{36.00} & \underline{33.00} & \underline{29.00} & 7.86 & 5.853 & 17.02 & 6.00 & \textbf{2.81} \\
RTC-Fixed & 25.00 & 31.00 & 21.00 & \textbf{0.48} & \textbf{0.345} & 11.41 & 5.18 & 4.02 \\
RTC-AHS & \textbf{39.00} & \textbf{39.00} & \textbf{34.00} & \underline{0.51} & \textbf{0.345} & 17.39 & 7.00 & \underline{2.90} \\
\midrule
\multicolumn{9}{l}{\textit{Place Can Basket}} \\
Sync-Fixed & 37.00 & 36.00 & 36.00 & 4.12 & \underline{4.832} & 14.05 & 5.75 & 5.09 \\
Sync-AHS & \textbf{47.00} & \underline{47.00} & \textbf{45.00} & 6.84 & 7.964 & 23.15 & 9.16 & \underline{3.12} \\
RTC-Fixed & \underline{39.00} & 37.00 & \underline{44.00} & \underline{0.33} & \textbf{0.344} & 13.83 & 7.17 & 5.06 \\
RTC-AHS & \textbf{47.00} & \textbf{50.00} & 39.00 & \textbf{0.30} & \textbf{0.344} & 26.57 & 12.60 & \textbf{3.00} \\
\bottomrule\end{tabular}\end{table}

\begin{figure}[!t]
  \centering
  \includegraphics[width=\linewidth]{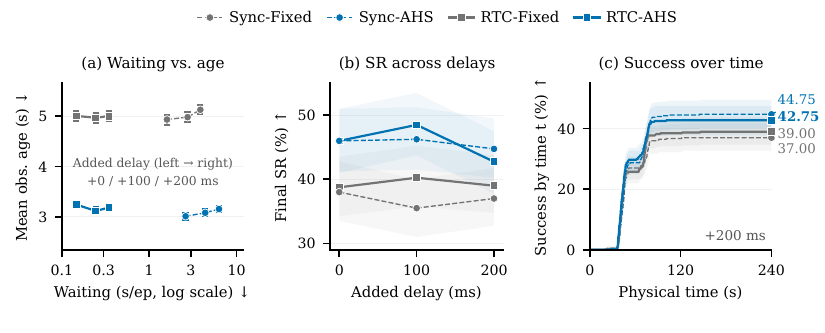}
  \caption{\textbf{AHS with asynchronous execution on $\pi_{0.5}$ Easy.}
  The three panels mirror Fig.~\ref{fig:rtc-latency}: waiting versus mean
  observation age at +0/+100/+200\,ms, final SR at each delay, and SR($t$)
  at +200\,ms. All 400 outcomes per condition are included. Bars and shading
  show marginal/pointwise 95\% paired-seed bootstrap intervals within four tasks.}
  \label{fig:rtc-latency-easy}
\end{figure}

\begin{table}[!t]\centering\small
\caption{\textbf{Aggregate costs at +0 and +100 ms on Easy and Hard.}
Each method/delay has 400 outcomes per setting, including failures.
The +200\,ms aggregates appear with the per-task results in
Tabs.~\ref{tab:rtc-full-panel} and~\ref{tab:rtc-easy-panel}.
Observation age averages within episodes, then equally across episodes and tasks.
Within each setting and delay, best and second-best distinct waiting and age values
are bold and underlined. Displayed ties share a mark.
Calls/ep and Infer s/ep are descriptive quantities and are not ranked.}
\label{tab:rtc-all-costs}\setlength{\tabcolsep}{3.2pt}
\begin{tabular}{@{}rlrrrrr@{}}\toprule
Delay (ms) & Method & Wait (\%) $\downarrow$ & Wait s/ep $\downarrow$ & Calls/ep & Infer s/ep & Mean age (s) $\downarrow$ \\
\midrule
\multicolumn{7}{l}{\textit{Easy}} \\
0 & Sync-Fixed & \underline{1.67} & \underline{1.600} & 11.11 & 4.32 & 4.93 \\
 & Sync-AHS & 2.87 & 2.614 & 18.15 & 6.78 & \textbf{3.01} \\
 & RTC-Fixed & \textbf{0.16} & \textbf{0.147} & 11.20 & 5.38 & 5.00 \\
 & RTC-AHS & \textbf{0.16} & \textbf{0.147} & 17.86 & 7.81 & \underline{3.24} \\
\midrule
100 & Sync-Fixed & 2.82 & \underline{2.755} & 11.29 & 4.37 & 4.98 \\
 & Sync-AHS & 4.80 & 4.384 & 17.97 & 6.60 & \textbf{3.08} \\
 & RTC-Fixed & \textbf{0.26} & \textbf{0.244} & 11.42 & 5.45 & 4.97 \\
 & RTC-AHS & \underline{0.28} & \textbf{0.244} & 18.73 & 8.09 & \underline{3.12} \\
\midrule
\midrule
\multicolumn{7}{l}{\textit{Hard}} \\
0 & Sync-Fixed & 1.60 & 1.853 & 12.87 & 5.14 & 4.84 \\
 & Sync-AHS & 2.93 & 3.244 & 22.53 & 8.37 & \textbf{2.85} \\
 & RTC-Fixed & \textbf{0.13} & \textbf{0.146} & 12.84 & 6.31 & 4.99 \\
 & RTC-AHS & \underline{0.14} & \underline{0.147} & 22.23 & 9.83 & \underline{3.01} \\
\midrule
100 & Sync-Fixed & \underline{2.72} & \underline{3.168} & 12.99 & 5.12 & 4.89 \\
 & Sync-AHS & 4.85 & 5.523 & 22.64 & 8.39 & \textbf{2.93} \\
 & RTC-Fixed & \textbf{0.22} & \textbf{0.244} & 13.34 & 6.64 & 4.84 \\
 & RTC-AHS & \textbf{0.22} & \textbf{0.244} & 24.05 & 10.49 & \underline{2.98} \\
\midrule
\bottomrule\end{tabular}\end{table}

\section{AHS Hyperparameter Sensitivity}
\label{sec:ahs-sensitivity}

We evaluate $\pi_{0.5}$ + AHS on eight RoboTwin2.0 tasks under Easy and
Hard settings, with $\mathcal K=\{10,20,30,40,50\}$ and expected-round
selection at $T_{\mathrm{sel}}=0.8$. The two studies below use different
episode budgets and pairing protocols. Neither changes our default configuration.

\begin{figure}[!t]
  \centering
  \includegraphics[width=\linewidth]{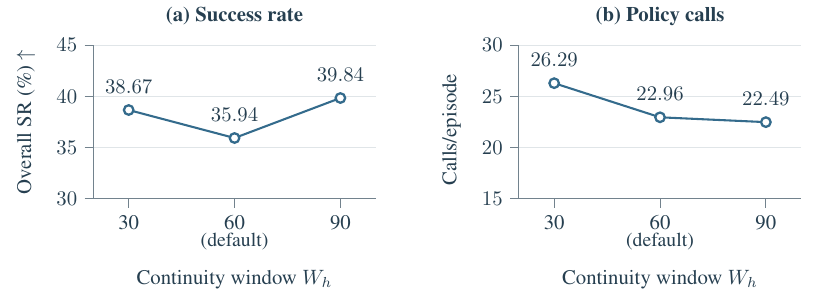}
  \caption{\textbf{Inter-chunk continuity-window sensitivity.}
  $\pi_{0.5}$ + AHS on eight RoboTwin2.0 tasks, Easy and Hard,
  with 256 matched episodes per configuration. Lines connect tested points.
  Paired difference intervals appear in the text.}
  \label{fig:continuity-window-sensitivity}
\end{figure}

\paragraph{Inter-chunk continuity window.}
We vary only $W_h\in\{30,60,90\}$, sharing 16 episode identities per
task and setting (256 per configuration). The default $W_h=60$ is
evaluated afresh as the paired reference in
Fig.~\ref{fig:continuity-window-sensitivity}.

Compared with $W_h=60$, gains for $30$ and $90$ are $+2.73$ and $+3.91$
percentage points, with paired 95\% confidence intervals
$[-2.34,\,7.81]$ and $[-0.78,\,8.59]$. We use 10,000 bootstrap draws
of matched episodes within task--setting cells, weighted equally.
Both intervals include zero, so neither a success-rate difference nor
equivalence is established. We retain $W_h=60$.

\paragraph{Other evidence and posterior hyperparameters.}
Each configuration in Tab.~\ref{tab:ahs-hyperparameter-sensitivity}
uses 100 episodes per task and setting (1,600 total), unpaired across
configurations. One parameter family changes at a time, with the two
penalties varied jointly. Other settings follow Appendix~\ref{sec:exp-setup}.
These evaluations are separate from the continuity-window study. Overall SR
spans 34.38--37.13\%, characterizing sensitivity rather than
validation-based selection or an established optimum.

\begin{table}[!t]
\centering
\caption{\textbf{Sensitivity to other AHS hyperparameters.}
Success rates (\%) on eight RoboTwin2.0 tasks, with 1,600 episodes per
configuration. Bold/underline mark higher/lower SR within each parameter family.
Defaults specify parameters, not additional evaluations.}
\label{tab:ahs-hyperparameter-sensitivity}
\begingroup
\fontsize{10}{11.5}\selectfont
\setlength{\tabcolsep}{3pt}
\renewcommand{\arraystretch}{1.02}
\begin{tabular*}{\linewidth}{@{\extracolsep{\fill}}lrrrrr@{}}
\toprule
Parameter & Default & Tested & Easy $\uparrow$ & Hard $\uparrow$ & Overall $\uparrow$\\
\midrule
Forgetting $\rho_f$ & 0.99 & 0.95 & \textbf{46.88} & \textbf{26.75} & \textbf{36.81}\\
 & & 0.995 & \underline{46.50} & \underline{25.88} & \underline{36.19}\\
\addlinespace[3pt]
Update strength $\eta$ & 1 & 0.5 & \underline{45.38} & \underline{24.38} & \underline{34.88}\\
 & & 2 & \textbf{47.38} & \textbf{26.88} & \textbf{37.13}\\
\addlinespace[3pt]
Kernel bandwidth $\sigma_K$ & 10 & 5 & \textbf{45.50} & \textbf{26.13} & \textbf{35.81}\\
 & & 20 & \underline{44.25} & \underline{24.50} & \underline{34.38}\\
\addlinespace[3pt]
Intra reference window $W_\tau$ & 3 & 1 & \textbf{47.38} & \textbf{24.75} & \textbf{36.06}\\
 & & 5 & \underline{46.38} & \underline{23.88} & \underline{35.13}\\
\addlinespace[3pt]
Joint $\alpha_{\mathrm{intra}}=\beta_{\mathrm{inter}}$ & 4 & 2 & \textbf{46.63} & \underline{23.63} & \underline{35.13}\\
 & & 8 & \underline{46.25} & \textbf{24.88} & \textbf{35.56}\\
\bottomrule
\end{tabular*}
\endgroup
\end{table}

\begin{figure}[!t]
  \centering
  \includegraphics[width=\textwidth]{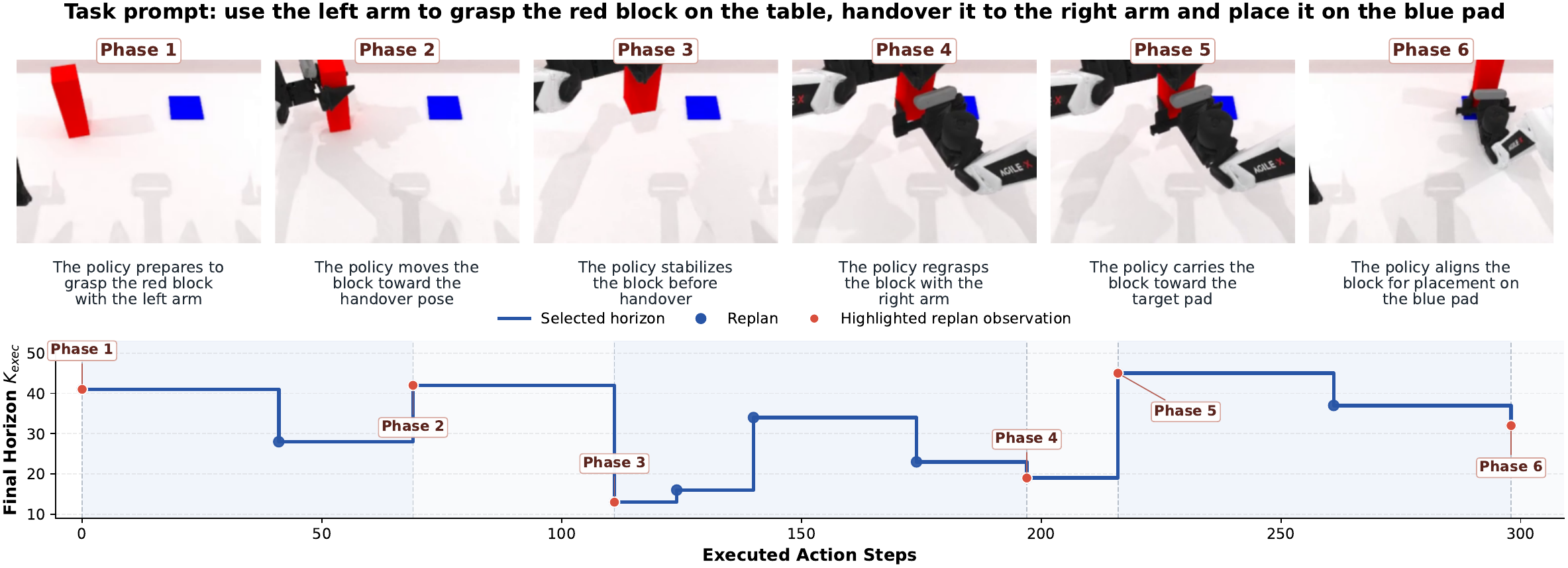}
  \caption{\textbf{$\pi_0$ Handover Block case study.}
  The selected execution horizon changes with task phase in a successful
  RoboTwin2.0 rollout. $K_{\mathrm{exec}}$ denotes the selected $K_t$.}
  \label{fig:handover-block-case-study}
\end{figure}

\section{More Case Studies}
\label{sec:more-case-studies}

\paragraph{Handover Block.}
Fig.~\ref{fig:handover-block-case-study} illustrates phase-dependent
execution in a successful $\pi_0$ rollout. AHS selects shorter horizons
during grasping and transport phases that require closed-loop correction,
and longer horizons once the motion stabilizes. This example illustrates
the behavior of the online selector in Sec.~\ref{subsec:test-time-method}.
It is not a controlled comparison of posterior mechanisms.

\begin{figure}[!t]
  \centering
  \includegraphics[width=\textwidth,height=0.90\textheight,keepaspectratio]{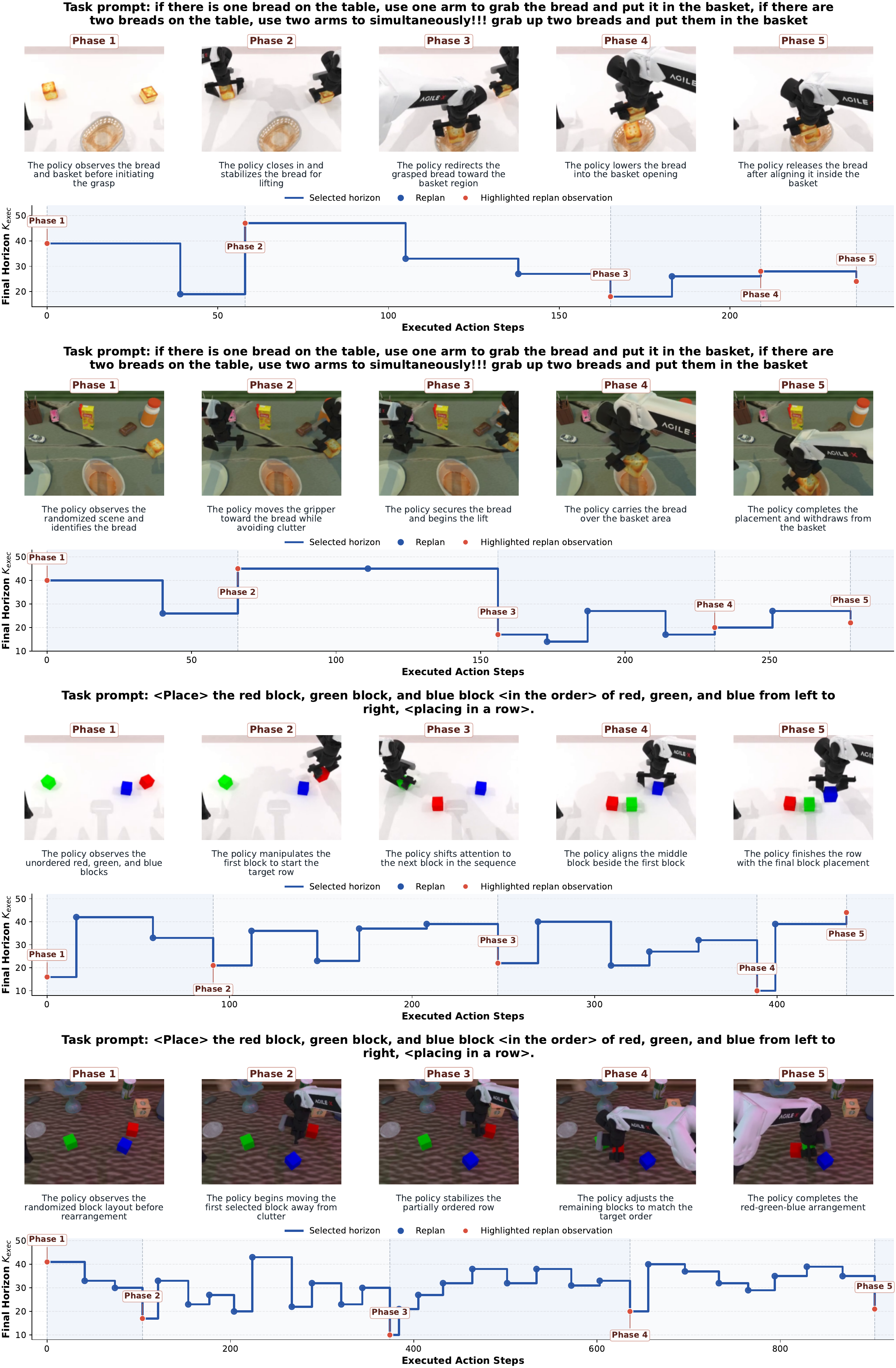}
  \caption{\textbf{Additional $\pi_0$ case studies on RoboTwin2.0.}
  Four successful rollouts show phase-dependent AHS horizons across
  Place Bread Basket and Blocks Ranking RGB under Easy and Hard settings.}
  \label{fig:app-more-case-studies}
\end{figure}

\begin{figure}[!t]
  \centering
  \includegraphics[width=\textwidth]{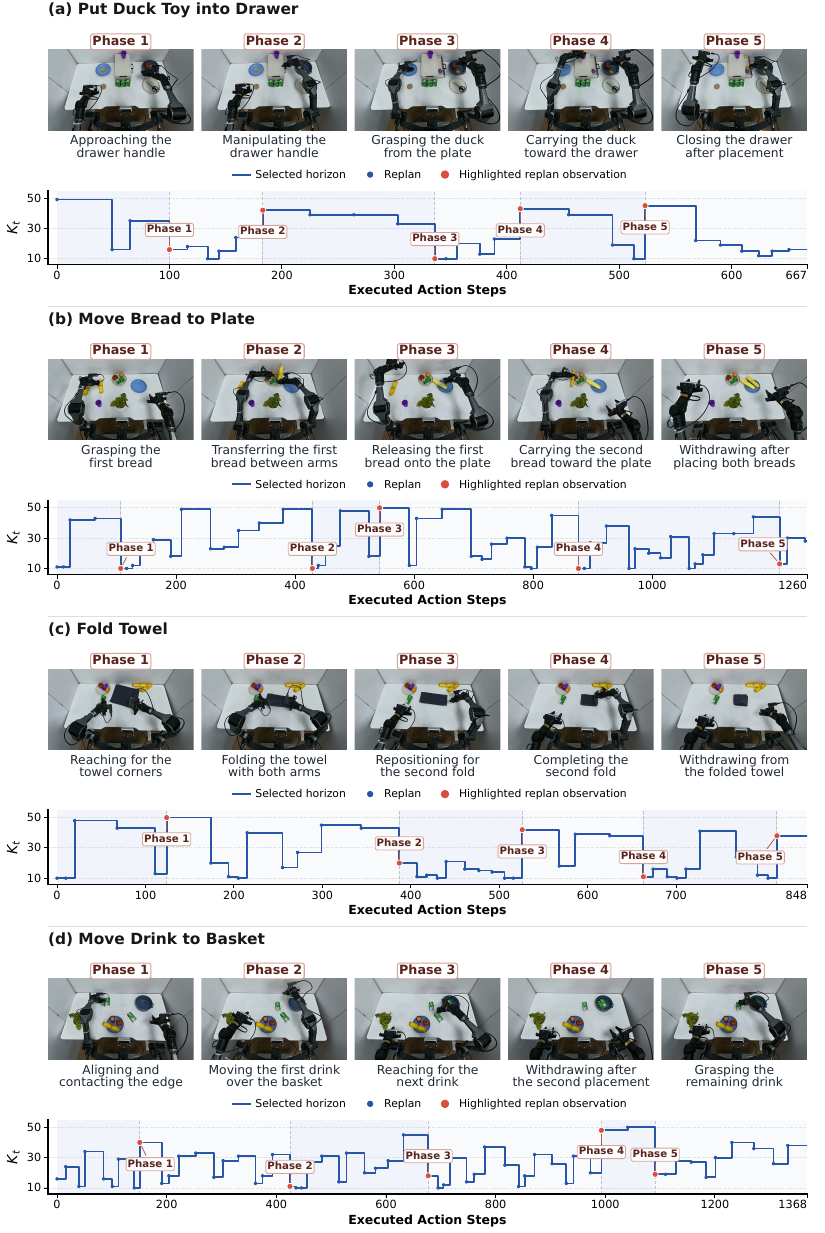}
  \caption{\textbf{Real-world $\pi_{0.5}$+AHS case studies.}
  Highlighted observations are selected at large adjacent-replan changes in
  $K_t$. Curves show all recorded execution horizons without smoothing,
  through the end of each recording. Phase labels are manual visual annotations, not
  ground-truth boundaries or trial scores.}
  \label{fig:app-real-world-case-studies}
\end{figure}

\paragraph{Additional tasks and settings.}
Fig.~\ref{fig:app-more-case-studies} shows four additional $\pi_0$
RoboTwin2.0 rollouts across Easy and Hard settings.
Fig.~\ref{fig:app-real-world-case-studies} adds $\pi_{0.5}$+AHS cases on
four real-world tasks. These qualitative examples illustrate horizon
changes across task phases, not additional scored trials.

\FloatBarrier
\section{Discussion and Limitations}
\label{sec:limitations-broader-impact}

\paragraph{Additional background.}
\label{sec:extended-related-work}
RT-1, RT-2, and PaLM-E connect large-scale learning with robot
control~\cite{brohan2022rt1,brohan2023rt2,driess2023palme}.
Open X-Embodiment, Octo, and OpenVLA extend shared data and generalist
initialization~\cite{openx2023,team2024octo,kim2024openvla}, while
RoboBrain 2.0 broadens embodied reasoning interfaces~\cite{team2025robobrain}.
RDT-1B and DexVLA extend diffusion-based action generation~\cite{liu2024rdt,wendexvla}.
Other approaches couple actions to world models~\cite{bi2025motus} or adapt
chunking through self-guidance~\cite{so2026improving}. ChunkTrust adapts
chunk execution using action-expert evidence.

\paragraph{Additional horizon and verification methods.}
Adaptive execution can draw on additional samples, learned sensitivity,
or observations acquired during rollout.
A3~\cite{chen2026a3} uses group-sampled consensus and conditional
re-decoding to verify a contiguous execution prefix.
SA~\cite{park2026spatial} forecasts the sensitivity of
the action distribution to observation changes and allocates shorter
horizons to more sensitive phases.
EQRL~\cite{wang2026eqrl} jointly learns the latent input, denoising budget,
and chunk length through reinforcement learning.
These methods differ in the information and computation used to choose
a horizon. AHS scores prefixes of one generated chunk using its existing
generation trace and executed history, with no conditional re-decoding or
joint optimization of the generator's inference schedule.
Verification methods instead use fresh observations to assess a running
plan. FFDC~\cite{wang2026ffdc} compares imagined futures with reality
for world-action models, while DREAM-Chunk~\cite{chen2026dreamchunk}
uses a latent world model to match candidate chunks' predicted futures
to observed execution.
SV-VLA~\cite{wang2026svvla} compares planned actions with a lightweight
closed-loop reference, and PATCH~\cite{zhou2026patch} accumulates
localized visual residuals along an action-conditioned execution corridor
to trigger intervention.
These observation-driven mechanisms address disturbances that become
visible after a chunk has been selected. ChunkTrust's evidence instead
informs how much of the current prediction to execute before observing
again, so it does not provide the same within-prefix monitoring capability.

\paragraph{Continuity and frequency-aware action generation.}
Cross-chunk consistency can be improved by changing how actions are
generated or corrected. ChunkFlow~\cite{yang2026chunkflow} trains
with seam and derivative-continuity losses and blends overlapping
predictions at execution. SEAM~\cite{zhan2026seam} steers denoising
toward the previous chunk's unexecuted tail, while
Legato~\cite{liu2026legato} learns continuation through action-conditioned
initialization and modified flow dynamics.
REMAC~\cite{wang2026remac} uses masked action conditioning for real-time
execution, and ACNet~\cite{guo2026acnet} conditions a lightweight
delay-aware adapter on executed motion.
A2C2~\cite{sendai2025a2c2} instead applies a learned per-step correction
using the latest observation and the base policy's action.
Frequency-aware approaches act on the representation or training objective.
FAFM~\cite{guo2026fafm} generates continuous action trajectories in a
frequency-domain representation, while FocalPolicy~\cite{he2026focalpolicy}
combines proximal time-domain supervision with multi-chunk spectral
regularization.
These works motivate attention to temporal coherence, but they do not
make the same intervention as ChunkTrust. Our spectral signal measures
variation during generation, and our continuity signal evaluates a prefix
stitched to executed history. Both are used to select an execution
length, leaving the generated action values and base-policy weights
unchanged. A frequency-domain training loss is therefore distinct from
the generation-time diagnostic used here.

\paragraph{Experience reuse and the role of chunking.}
TraceFlow~\cite{zhang2026traceflow} reuses successful and failed rollouts
through a retrieval bank and outcome-conditioned guidance of a frozen
flow-matching action expert. ChunkTrust reuses information in a different
form and for a different decision. AHS retains evidence-derived horizon
preferences within an episode, while QHA learns a context-conditioned
horizon prior offline. Neither component retrieves rollout trajectories
to steer action generation, and QHA does not update its weights during
deployment.
Recent analyses also caution against treating long open-loop execution
as universally beneficial. \citet{lazzati2026chunking}
study non-Markovian expressivity and implicit ensembling as explanations
for the benefits of chunking.
\citet{zeng2026openloop} show that the value of open-loop
execution depends on demonstration non-Markovianity and policy context
length. ChunkTrust addresses horizon selection for existing chunk policies,
not a claim that longer open-loop execution is intrinsically preferable
to reactive control.

\paragraph{Limitations.}
AHS requires access to action chunks and generation-time velocity traces.
It scores a candidate grid, although expected-round selection can return
intermediate integer lengths. QHA learns a dense prior, but AHS+QHA still
computes online evidence, interpolated or scored directly on the dense grid.
Horizon decisions are made at replanning time, so the selector cannot
directly detect a new disturbance that arises during the chosen prefix.
Our evaluations cover multiple policies and two simulation benchmarks,
plus four real-world bimanual tasks, rather than all embodiments,
safety-critical tasks, or long-horizon mobile manipulation.

\paragraph{Broader impact.}
Adaptive horizons may reduce unnecessary replanning while preserving
reactivity near contact. An incorrect horizon can still commit a robot
to unsafe motion outside the tested distribution. Deployment should retain
workspace and speed limits, emergency stops, human supervision during
evaluation, and task-specific validation.

\paragraph{Third-party resources.}
\label{sec:existing-assets}
We use the cited RoboTwin2.0 and RoboCasa benchmarks, policy checkpoints,
and associated software for research evaluation. These third-party resources
remain subject to their original licenses, terms of use, and attribution
requirements.

\end{document}